\documentclass{article}

\usepackage{microtype}
\usepackage{graphicx}
\usepackage{subcaption}
\usepackage{booktabs} 

\usepackage{hyperref}

\usepackage[accepted]{icml2026}

\usepackage{amsmath}
\usepackage{amssymb}
\usepackage{mathtools}
\usepackage{amsthm}

\usepackage[capitalize,noabbrev]{cleveref}

\usepackage[utf8]{inputenc} 
\usepackage[ruled,vlined]{algorithm2e}
\usepackage{listings}
\usepackage{tcolorbox}
\usepackage[dvipsnames,table]{xcolor}
\usepackage{textgreek}
\usepackage{enumitem}
\usepackage{xspace}
\usepackage{csquotes}
\usepackage{etoolbox}

\MakeOuterQuote{"}

\definecolor{LightBlue}{rgb}{0.9, 0.96, 1}

\definecolor{thinkcolor}{RGB}{100,149,237}
\definecolor{answercolor}{RGB}{34,139,34}

\makeatletter
\lst@Key{escapeinside}{}{\def\lst@escapeinside{#1}}
\makeatother

\newtcolorbox{promptbox}{
  colback=gray!10,
  colframe=gray!40,
  arc=4mm,
  boxrule=0pt,
  left=5pt, right=5pt,
  top=5pt, bottom=5pt,
  fontupper=\footnotesize
}

\newcommand{\commenttt}[1]{\textcolor{gray}{\textit{#1}}} 

\newcommand{\think}[1]{\textcolor{thinkcolor}{#1}}
\newcommand{\answer}[1]{\textcolor{answercolor}{#1}}

\newcommand{\showbest}{Best performance in \textbf{bold}.\xspace{}}

\usepackage{pifont}

\usepackage{microtype}

\renewcommand{\paragraph}[1]{\vspace{.5em}\noindent\textbf{#1.}}

\AfterEndEnvironment{tabularx}{\vspace{-2em}}
\AtEndEnvironment{figure}{\vspace{-1em}}

\usepackage[utf8]{inputenc} 
\usepackage[T1]{fontenc}    
\usepackage{url}            
\usepackage{amsfonts}       
\usepackage{nicefrac}       
\usepackage{xcolor}         
\usepackage{xspace}
\usepackage{multicol}
\usepackage{multirow}
\usepackage{tabularx}
\usepackage{bm}
\usepackage{caption}
\usepackage{threeparttable}
\usepackage[normalem]{ulem}
\useunder{\uline}{\ul}{}
\usepackage{placeins}
\usepackage{array}
\usepackage{tikz}
\usetikzlibrary{positioning}
\usepackage{wrapfig,lipsum}

\newcommand{\yangcx}[1]{}

\newcolumntype{P}[1]{>{\centering\arraybackslash}p{#1}}
\newcolumntype{Y}{>{\centering\arraybackslash}X}
\newcommand{\methodName}{APEIRIA\xspace}

\newcommand{\hideold}[1]{}

\renewcommand{\cite}{\citep}

\theoremstyle{plain}

\theoremstyle{definition}

\theoremstyle{remark}

\icmltitlerunning{Distilling Neuro-Symbolic Programs into 3D Multi-modal LLMs}

\begin{document}

\twocolumn[
  \icmltitle{Distilling Neuro-Symbolic Programs into 3D Multi-modal LLMs}

  \icmlsetsymbol{equal}{*}

  \begin{icmlauthorlist}
    \icmlauthor{Wentao Mo}{yyy}
    \icmlauthor{Yang Liu}{yyy}
  \end{icmlauthorlist}

  \icmlaffiliation{yyy}{Wangxuan Institute of Computer Technology, Peking University}

  \icmlcorrespondingauthor{Yang Liu}{yangliu@pku.edu.cn}

  \icmlkeywords{Neuro-Symbolic, 3D LLM, Multimodal, ICML}

  \vskip 0.3in
]

\printAffiliationsAndNotice{}  

\begin{abstract}
Current 3D spatial reasoning methods face a fundamental trade-off: neuro-symbolic 3D (NS3D) concept learners achieve interpretable reasoning through compositional programs but are constrained to closed-set concept vocabularies and simple programs; end-to-end 3D multi-modal LLMs (3D MLLMs) could handle complex natural language and open-vocabulary concepts but suffer from black-box reasoning without explicit spatial verification.
We introduce \methodName{}, a neuro-symbolic 3D MLLM to bridge two paradigms by distilling symbolic reasoning patterns into MLLMs with natural language chain-of-thought.
Our three-stage curriculum progressively builds reasoning capabilities:
a) 3D perception alignment grounds object visual-geometric features to the LLM,
b) CoT-SFT teaches query decomposition and stepwise verification from symbolic program traces,
and c) CoT-RL extends reasoning patterns to open-set concepts and deeply nested instructions.
By transferring reasoning patterns rather than concept-specific knowledge, \methodName{} preserves key NS3D virtues: transparent reasoning and modular interchangeability of planning and perception components.
Evaluations on grounding, question answering, and captioning show that \methodName{} surpasses prior NS3D methods and matches state-of-the-art 3D MLLMs on 3D spatial reasoning datasets, unifying symbolic methods' systematic reasoning with MLLMs' flexibility.
Code is available at \url{https://github.com/oceanflowlab/APEIRIA}.

\end{abstract}

\section{Introduction}
\label{sec:intro}

\begin{figure}[!ht]
    \centering
    \includegraphics[width=\linewidth]{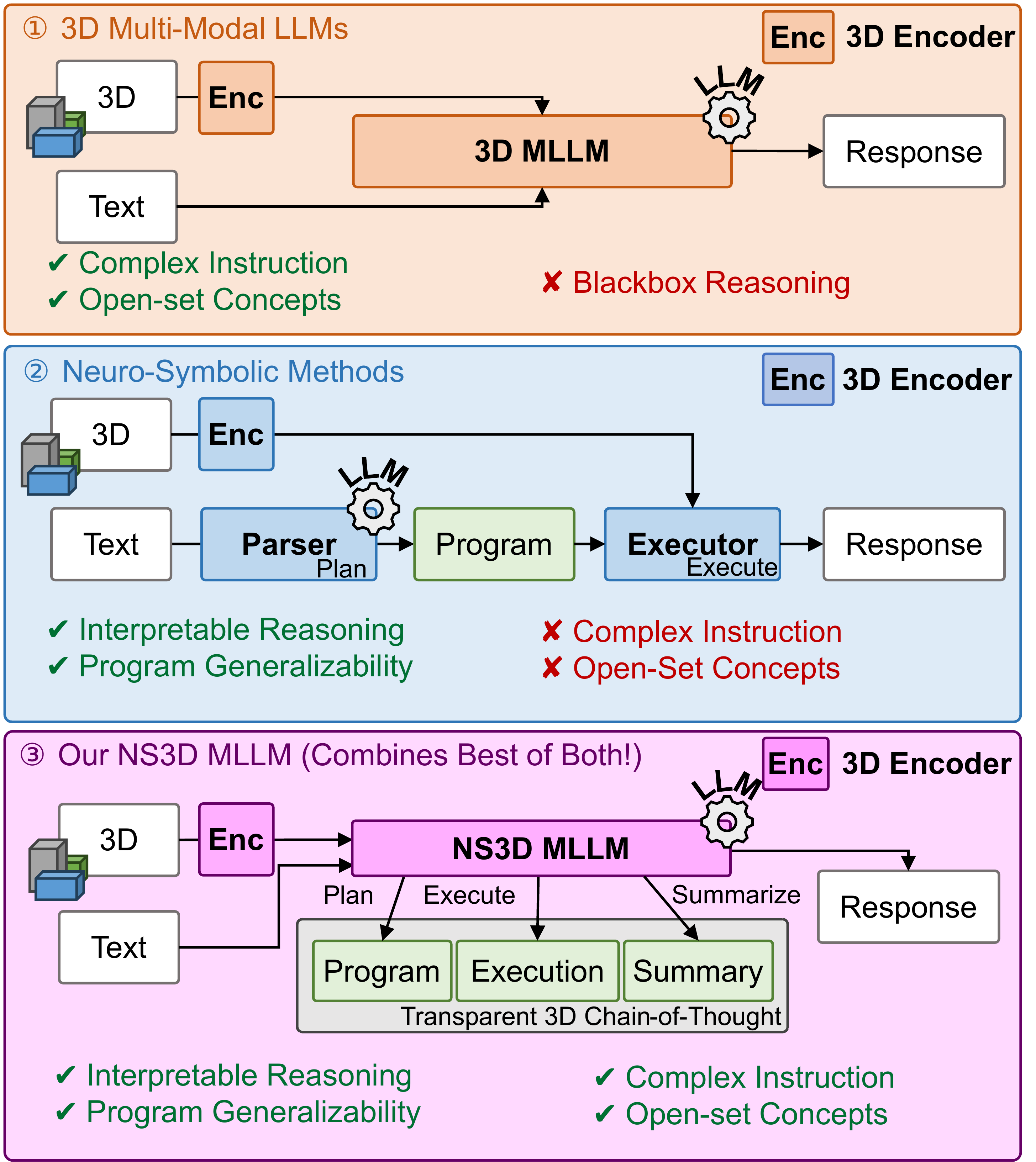}
    \caption{
    \textbf{Bridging 3D MLLMs and neuro-symbolic reasoning.}
    \ding{192} 3D multi-modal LLMs could handle complex natural language and open-vocabulary instructions but reason as black boxes without interpretable verification.
    \ding{193} Neuro-symbolic 3D (NS3D) methods offer transparent, step-by-step reasoning programs but are limited to closed-set concepts and require dense procedure supervision.
    \ding{194} Our method bridges them by distilling symbolic reasoning patterns 
    from programs into transparent chain-of-thought within 
    3D MLLM, combining the best of both methods.
    }
    \label{fig:teasor}
\end{figure}

3D spatial reasoning is fundamental to embodied AI and 3D scene understanding, and two paradigms have emerged to tackle this challenge (\Cref{fig:teasor}).
The dominant paradigm, \textbf{3D Multi-modal LLMs (3D MLLMs)}~\cite{chen2023ll3da,huang2023embodied,huang2024chat,zhu2025llava3dsimpleeffectivepathway,zheng2024video3dllmlearningpositionaware,Inst3D-LMM,chen2024grounded3dllmreferenttokens} (\Cref{fig:teasor}\ding{192}), leverages LLMs' semantic power to handle free-form referring expressions.
Yet, lacking explicit reasoning structure, these \textbf{black-box} models directly map instructions to answers without interpretable verification steps. When they fail, there is no trace to diagnose \textit{where} the error occurred in object recognition, relation understanding, or logical composition.
In contrast, a smaller line of work on \textbf{Neuro-Symbolic 3D (NS3D) Learners}~\cite{Hsu2023NS3DNG, Feng2024NaturallyS3, Yi2018NeuralSymbolicVD} (\Cref{fig:teasor}\ding{193}) pursues interpretable spatial reasoning by decomposing queries into modular programs iteratively executed by trained concept-specific networks. 
While excelling at systematic generalization within closed-set vocabularies, they face two fundamental barriers: 
(\textbf{i}) rigid concept-specific networks cannot handle \textbf{open-vocabulary} spatial queries (e.g., ``cozy chair'', ``messy desk''); 
(\textbf{ii}) training modular components requires \textbf{dense step-by-step supervision} unavailable beyond simple synthetic datasets.

The field appears trapped in a trade-off: interpretable but limited NS3D learners versus powerful but opaque MLLMs. 
However, we identify an opportunity to decouple \textbf{concept learning} from \textbf{reasoning learning}. Neuro-symbolic programs encode precise, systematic reasoning patterns (the ``syntax'' of thought) which can be distilled into the MLLM, while the MLLM itself provides the open-world knowledge to understand 3D semantics.
On synthetic datasets like Sr3D~\cite{achlioptas2020referit_3d} where instructions are template-generated from heuristics, the symbolic programs are guaranteed to have \textit{complete intermediate supervision}: every primitive's input and output can be constructed and verified against ground-truth annotations, which is a supervision often unavailable in real-world datasets. This provides an ideal stepping stone for transferring reasoning capabilities into 3D MLLMs: we could first distill these verified execution traces into MLLMs, teaching the model systematic query decomposition and stepwise spatial verification skills. Then, on real-world datasets where such complete traces are unavailable, we employ reinforcement learning with outcome supervision to bootstrap and extend these learned reasoning patterns to open-vocabulary concepts and complex nested instructions.

Building on this insight, we present \methodName{}\footnote{\textAlpha\textpi\textepsilon\textiota\textrho\textiota\textalpha, \textit{Unlimited} in Greek.}, a neuro-symbolic MLLM for 3D spatial reasoning, that unifies the systematic rigor of symbolic methods with the semantic flexibility of LLMs (\Cref{fig:teasor}\ding{194}). 
Built upon an efficient object-centric representation, \methodName{} is trained through a \textbf{Curriculum-based Reasoning Distillation} framework that progressively builds spatial reasoning capabilities.
In \textbf{Stage 1 (Perception Alignment)}, we teach the model to \textit{see}, by aligning 3D visual-geometric features with the LLM's embedding space to recognize object categories, attributes, and locations.
In \textbf{Stage 2 (Symbolic Reasoning Injection)}, we teach the model to \textit{think iteratively}, by learning from \textbf{verified reasoning traces} from symbolic programs via CoT-SFT, where each trace contains explicit plans (query decomposition) and executions (spatial verification with \textbf{object IDs and locations}), providing \textbf{correct} supervision for systematic spatial reasoning.
In \textbf{Stage 3 (Open-Set and Complex Reasoning Generalization)}, we teach the model to \textit{adapt} to real-world instructions, by using CoT-RL to extend these learned patterns to \textbf{open-vocabulary concepts} and \textbf{complex nested queries} where step-by-step supervision lacks.
This yields an end-to-end model that solves complex 3D spatial reasoning tasks while retaining key neuro-symbolic virtues: \textbf{inference transparency} through explicit reasoning traces, and \textbf{modularity} of planning and perception components interchangeable as stronger LLMs or 3D perception models emerge.

Concretely, our key contributions are: 
1) We introduce \methodName{}, a \textbf{neuro-symbolic 3D MLLM} that bridges symbolic reasoning and modern MLLMs. By explicitly modeling reasoning processes as transparent chain-of-thought, it enables systematic, generalizable planning for 3D spatial reasoning tasks.
2) We propose a \textbf{curriculum-based reasoning distillation} framework that progressively transfers reasoning patterns from neuro-symbolic programs into 3D MLLMs: from perception alignment, through symbolic reasoning injection, to open-set generalization via reinforcement learning.
3) We demonstrate that \methodName{} surpasses previous neuro-symbolic 3D methods and state-of-the-art 3D MLLMs on 3D spatial reasoning benchmarks, while maintaining transparent reasoning and efficient generalization without excessive tokens or auxiliary architectures.

\section{Related Work}

\noindent \textbf{Neuro-Symbolic 3D (NS3D) Grounding and Reasoning.} 
Neuro-symbolic methods~\cite{Yi2018NeuralSymbolicVD, johnson2017inferringexecutingprogramsvisual, vedantam2019probabilisticneuralsymbolicmodelsinterpretable,mao2019neurosymbolicconceptlearnerinterpreting} have demonstrated exceptional data efficiency and generalization in the 2D visual reasoning. 
In the 3D domain, approaches~\cite{Hsu2023NS3DNG, Feng2024NaturallyS3} decompose spatial queries into compositional programs executed by concept-specific neural modules, enabling zero-shot transfer to novel program structures.
However, NS3D methods face two limitations:
\textbf{(i)} their reliance on concept-specific networks restricts them to closed-set vocabularies: a "chair" detector can't recognize "cozy furniture";
\textbf{(ii)} training modular components requires \textbf{step-by-step supervision} only available in simple synthetic datasets like Sr3D~\cite{achlioptas2020referit_3d} with at most 2-level nested programs, preventing scaling to complex real-world instructions.
Our method addresses these gaps by distilling reasoning patterns into MLLMs for open-vocabulary understanding, and is the first to introduce outcome-based RL in NS3D methods to extend NS3D methods beyond procedural supervision requirements.

\noindent \textbf{3D Multi-Modal Large Language Models (MLLMs) for Spatial Reasoning.} 
There has been a surge of interest in integrating LLM with 3D spatial understanding capabilities ~\cite{chen2023ll3da,huang2024chat,Mo_Liu_2024,zhu2025llava3dsimpleeffectivepathway,zheng2024video3dllmlearningpositionaware,Inst3D-LMM,chen2024grounded3dllmreferenttokens,mo2025lego,Ground_2025_ICRA,zhou2025aqua,zhao2025domaingaps-arxiv,zhou2026scalableobjectrelationencoding}. 
These methods range from token-efficient object-centric approaches  ($<$1K scene tokens) ~\cite{huang2024chat,Inst3D-LMM,chen2023ll3da} to video-centric methods~\cite{zhu2025llava3dsimpleeffectivepathway,zheng2024video3dllmlearningpositionaware} (10K--40K scene tokens), yet both treat reasoning as a \textbf{black-box mapping} without explicit verification steps.
When grounding fails, there is no reasoning trace to diagnose whether the error lies in object recognition, relation understanding, or logical composition.
Our work aligns with the streamlined object-centric architecture for efficiency, but uniquely injects \textbf{structured, interpretable reasoning traces} that expose the model's spatial verification process, enabling transparent debugging and modular component replacement.

\noindent \textbf{Reinforcement Learning (RL) for Visual Reasoning.} 
Recently, RL techniques with chain-of-thought have recently been introduced to enhance MLLM reasoning beyond supervised fine-tuning and beyond supervised fine-tuning on black-box answer generation, with notable progress in 2D vision-language understanding and generation tasks~\cite{zhang2025r1,peng2025lmmr1empowering3blmms,Qin2025ApplyHT,peng2025,lyu2025, Gong2026}.
In the 3D domain, concurrent works 3D-R1~\cite{huang20253dr1enhancingreasoning3d} and Scene-R1~\cite{yuan2025scener1videogroundedlargelanguage} represent initial attempts to apply RL for 3D spatial reasoning. 
However, these 3D approaches have critical limitations.
Scene-R1 lacks any reasoning trace enforcement or initialization, making RL exploration unstable and inefficient.
3D-R1 generates CoT via LLM prompting, which is prone to hallucination and produces traces without \textbf{explicit spatial grounding}: objects are referenced vaguely rather than by precise IDs, locations, or bounding boxes, limiting utility for spatial verification.
In contrast, our approach first distills \textbf{verified, spatially-grounded traces} from symbolic programs, where each step explicitly references objects by ID and spatial configuration.
This structured initialization enables RL to refine and extend reasoning patterns rather than discover them from scratch, yielding more robust and interpretable spatial reasoning.

\section{\methodName: Progressive Distillation from Symbolic Programs to Interpretable 3D Reasoning}
\label{sec:method}

\begin{figure*}
    \centering
    \includegraphics[width=\linewidth]{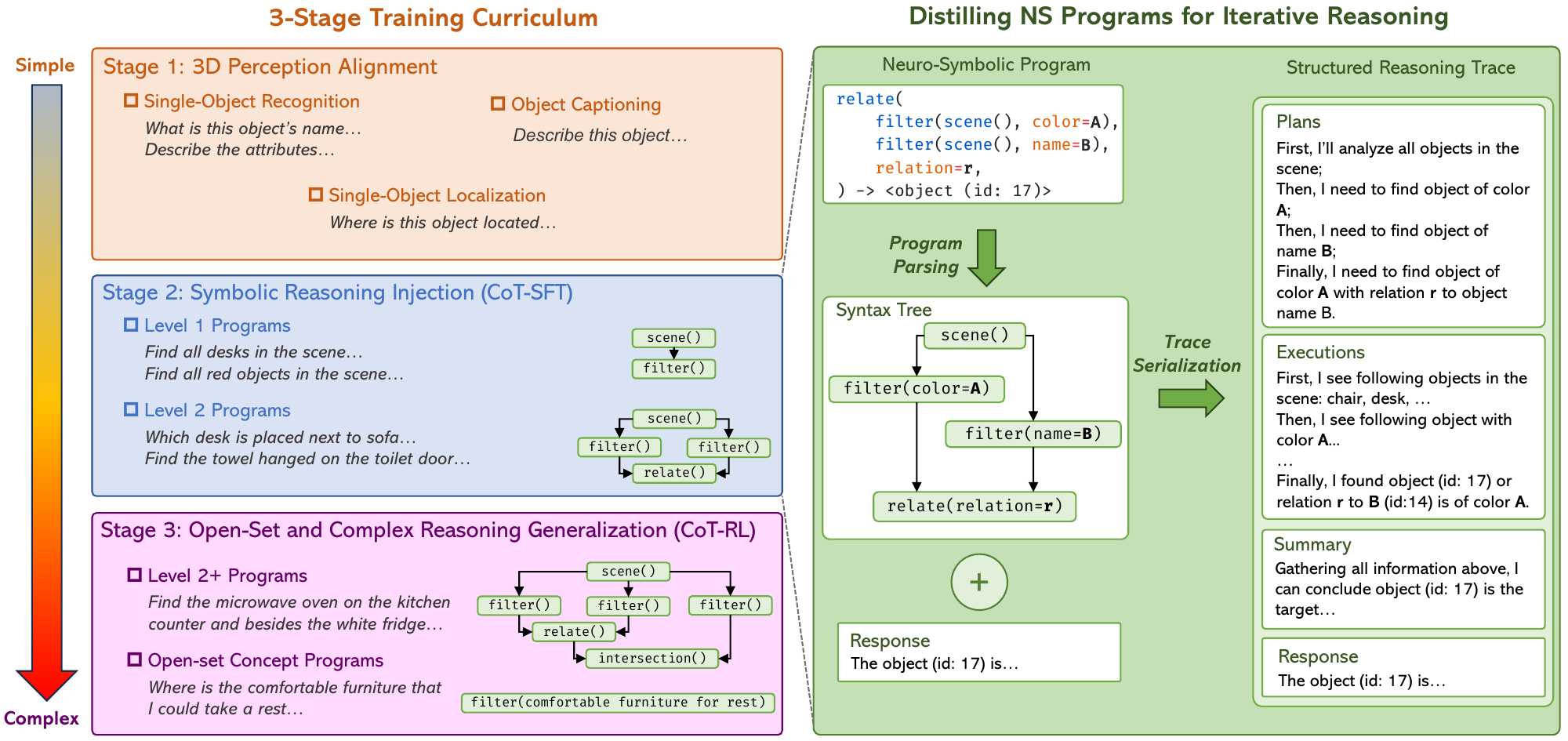}
\caption{
\textbf{Progressive Curriculum-based Reasoning Distillation Framework.}
\textbf{Left: Three-Stage Simple-to-Complex Curriculum.}
\ding{192} \textbf{Perception Alignment} grounds 3D visual features to textual concepts.
\ding{193} \textbf{Symbolic Reasoning Injection} distills reasoning patterns from programs via CoT-SFT, providing procedural supervision for query decomposition and execution.
\ding{194} \textbf{Open-Set and Complex Reasoning Generalization} extends these patterns to complex, open-vocabulary scenarios via CoT-RL with outcome supervision.
\textbf{Right: Program-to-CoT Translation.}
To generate Stage \ding{193} supervision, we parse neuro-symbolic programs into execution traces, which are then \textbf{serialized} into natural language CoT. This explicitly maps symbolic operations to transparent reasoning steps (plans and executions), enabling the MLLM to learn the "syntax" of spatial reasoning.
}
\label{fig:data}
\end{figure*}

\subsection{Object-Centric 3D Representation}
\label{sec:3d-repr}

To equip \methodName with a comprehensive understanding of the 3D environment, we first encode both the visual properties of object instances and their spatial configurations.
We adopt an object-centric scene representation that enables efficient reasoning with minimal input tokens ($\approx$400 tokens vs. 10k-40k in video-based MLLMs~\cite{zheng2024video3dllmlearningpositionaware,zhu2025llava3dsimpleeffectivepathway}).
Following prior work~\cite{huang2024chat}, the input 3D scene is first segmented into object instances using Mask3D~\cite{Schult23ICRA}.
For each instance, we extract 3D visual-geometric features with Uni3D~\cite{Zhou2023Uni3DEU} and 2D appearance features with DINOv2~\cite{Oquab2023DINOv2LR}.
Spatial information (object location and dimensions) is encoded via learnable positional embeddings.
The final object representation concatenates visual and spatial features, which are interleaved with instruction tokens as LLM input.
This compact representation explicitly preserves object-level structure, while maintaining facilitating architectural simplicity. Please refer to the supplementary material for detailed implementation.

\subsection{Curriculum-based Reasoning Distillation}
\label{sec:distillation}

Our core insight is that neuro-symbolic programs encode the exact reasoning patterns we wish to preserve—systematic decomposition, explicit intermediate states, and compositional execution—making them ideal supervision signals for teaching 3D MLLMs interpretable reasoning.
Therefore, we propose to \textit{distill the reasoning patterns} from NS programs into natural language chain-of-thought into 3D MLLMs. This approach eliminates the dependency on rigid, concept-specific networks, such as the separate networks for each object category (e.g., $f_{\text{chair}}$) or relation (e.g., $f_{\text{left}}$) required by traditional NS3D learners~\cite{Hsu2023NS3DNG}, thereby allowing us to leverage the open-vocabulary semantic flexibility of LLMs while preserving systematic generalization.

This distillation proceeds through a progressive three-stage curriculum (\Cref{fig:data}), transforming the model to an active reasoner:
\textbf{Stage 1: Perception Alignment (\Cref{sec:alignment}).} We first teach the model to \textit{see} by aligning 3D visual-geometric features with LLM's textual representation space, establishing the foundational ability to recognize and locate objects.
\textbf{Stage 2: Symbolic Reasoning Injection (\Cref{sec:cot-sft}).} We then teach the model to \textit{think} interatively by distilling reasoning traces from symbolic programs. Through CoT-SFT, the model learns the "syntax" of spatial reasoning: how to decompose queries and verify them step-by-step.
\textbf{Stage 3: Open-Set and Complex Reasoning Generalization (\Cref{sec:cot-rl}).} Finally, we teach the model to \textit{adapt} to real-world scenarios. Using CoT-RL, we bootstrap the learned reasoning patterns to handle open-vocabulary concepts and complex nested instructions where procedural supervision for each step is unavailable.
This curriculum efficiently bridges the gap between limited high-quality symbolic supervision and broad, real-world reasoning capabilities.

\subsubsection{Perception Alignment}
\label{sec:alignment}

Prior to program distillation, we align 3D object representations with the LLM's textual embedding space through standard 3D vision-language pre-training~\cite{huang2023embodied, zhu2025llava3dsimpleeffectivepathway, mo2025lego}.
This stage trains on mixture of object-centric perception tasks: \textbf{object recognition} (identifying categories/attributes), \textbf{localization} (predicting coordinates), and \textbf{captioning} (generating descriptions),
totaling approximately 193K instruction-response pairs (please refer to~\Cref{sec:data_details} for data details).
This establishes a robust foundation for 3D scene understanding, enabling the model to ground basic object concepts before tackling complex compositional reasoning.

\subsubsection{Symbolic Reasoning Injection}
\label{sec:cot-sft}

Having established basic 3D understanding capabilities, we now inject systematic reasoning patterns from neuro-symbolic programs with Chain-of-Thought Supervised Fine-Tuning (CoT-SFT).
NS programs offer a `white-box' supervision signal: every execution step provides ground-truth inputs and outputs. This allows us to teach the MLLM \textit{how to think} (plan decomposition)—how to decompose and execute complex queries—rather than just what to answer.
In addition, this pattern-level transfer is crucial for subsequent generalization to novel concepts and structures in Stage 3 (\Cref{sec:cot-rl}).

\noindent\textbf{Neuro-Symbolic Programs as Supervision.}
We adopt the neuro-symbolic framework from NS3D~\cite{Hsu2023NS3DNG}, which decomposes visual reasoning instructions into compositional programs comprising primitive operations. Each primitive accepts object sets and conditions as inputs and return filtered or related object sets.
Specifically, \lstinline|scene()| initializes and returns the set with all scene objects $\mathcal{O}$;
\lstinline|filter($\mathcal{O}$,condition)| selects objects matching specific semantic attributes (e.g., category or color);
\footnote{For brevity, we denote \lstinline|filter(scene(),condition)| as \lstinline|filter(condition)|.}
and \lstinline|relate($\mathcal{O}_A$,$\mathcal{O}_B$,r)| returns objects in $\mathcal{O}_A$ with spatial relation $r$ to any object in $\mathcal{O}_B$, while \lstinline|relate_triple| extends this to triplet relations.
These primitives compose into programs for complex queries, e.g., \lstinline|relate(filter(desk), filter(wall), left)| finds desks left of walls.
Crucially, while traditional NS3D methods implement these primitives via rigid concept-specific networks with heuristic designs that hinder open-set generalization, we propose to leverage the semantic flexibility of MLLMs to execute each primitive through natural language reasoning. To achieve this, we construct a corpus of plan-then-execution CoT to teach model how to reason systematically.

\noindent \textbf{Constructing Programs with Complete Execution Supervision.}
We leverage existing annotated 3D datasets to build a diverse set of programs with \textbf{complete execution traces}: every intermediate execution step have verified output from ground-truth annotations.

\noindent \textit{Level 1 (Single-step programs):}
Instructions involve only single \lstinline|filter| operations, such as "find all desks" (\lstinline|filter(desk)|) or "find red objects" (\lstinline|filter(red)|).
we utilize category and attribute annotations from ScanNet~\cite{dai2017scannet} and MMScan~\cite{mmscan} to provide ground-truth outputs for each \lstinline|filter| operation, teaching the model basic visual-geometric semantics.

\noindent \textit{Level 2 (Two-step programs):}
For relational reasoning, we utilize the Sr3D dataset~\cite{achlioptas2020referit_3d}, which contains synthetic instructions guaranteed to be solvable in exactly two steps.
Instructions in Sr3D follow these fixed compositional patterns, and can be directly mapped to two-step programs:
(\textbf{i}) "find A with relation r to B" (\lstinline|relate(filter(A),filter(B),r)|);
(\textbf{ii}) "find A with relation r to B and C" or "find A with relation r with B, facing C" (\lstinline|relate_triple(filter(A),filter(B),filter(C),r)|).
The synthetic nature of Sr3D allows us to derive complete intermediate supervision: the outputs of \lstinline|filter| steps are verified by object annotations, while the final \lstinline|relate| outputs are verified by target annotations.
In total, we construct 78K Level-1 traces from ScanNet and MMScan annotations and 66K Level-2 traces from Sr3D, yielding 144K verified CoT examples for this stage.

However, this reliance on perfect intermediate supervision limits scalability to real-world datasets (e.g., ScanRefer), where ground-truth execution procedural supervision are \textbf{unavailable for instructions corresponding to level 2+ programs}, necessitating the adaptive reinforcement learning introduced in Stage 3 (\Cref{sec:cot-rl}).

\begin{figure}[!t]
\centering
\begin{promptbox}
These are all objects in the scene: |object\_set|. Find the vase left to the computer.

\think{[think]}

I need to find the vase left to the computer.

\commenttt{// Stepwise plans.}

Let's plan my next steps: Examine all objects in the scene; Find all objects of category 'vase'; Find all objects of category 'computer'; Check which vase(s) are 'left to' the computer.

\commenttt{// Execution of scene() statement.}

First, let me list the objects and names:
Object 0(vase), 1(vase), 2(bottle), 3(rug), ... \textcolor{gray}{[others omitted]}

\commenttt{// Execution of filter() statement.}

Among these objects, I can see the vase(s):

Object 0: At (6.41,...), size:1.60 x...\textcolor{gray}{[more objects omitted]}


And the computer(s):

Object 4: At (1.44,...), size:0.87 x...\textcolor{gray}{[more objects omitted]}


\commenttt{// Execution of relate() statement.}

Now, I will analyze if any vase is left to the computer:

The vase (ID 0) is left to the computer (ID 4); ...\textcolor{gray}{[more relations omitted]}




\commenttt{// Summarization.}

From analysis above, the vase (ID 0) is left to a computer (ID 4).
Now, I will formulate the response based on the identified objects.

\answer{[answer]}

I found the vase left to the computer:
Object 0: At (1.15, 6.09, 1.33), size: 0.86 x 0.99 x 1.79
\end{promptbox}
\caption{Example CoT for a \lstinline|relate| program.}
\label{fig:cot_example}
\end{figure}

\noindent \textbf{Composing Chain-of-Thought from Programs.}
As illustrated in \Cref{fig:data} (right), we \textbf{translate program into natural language CoT} through parsing and serialization.
For each program, we parse its abstract syntax tree into an execution sequence $\mathcal{S} = \{s_1, \ldots, s_n\}$.
Each step $s_i$ is serialized into natural language via two components:
(\textbf{i}) a \textit{plan} describing the reasoning goal (e.g., ``Find all objects of category `vase' ''), and
(\textbf{ii}) an \textit{execution} presenting the step's inputs and outputs with object details (IDs, positions, sizes).
We use predefined templates that fill in ground-truth object sets at each step.
The final CoT concatenates all plans followed by all executions, creating a transparent reasoning trace from query to answer (\Cref{fig:cot_example}).
Crucially, our CoT is \textbf{spatially grounded}: each object is referenced by a unique ID with explicit locations, enabling disambiguation of same-category objects and precise spatial verification.
The details are provided in Algorithm~\ref{alg:build_cot} in the supplementary material.

\noindent \textbf{Training Objective.}
We train \methodName's parameters $\theta$ to generate CoT traces via standard language modeling:
{
\setlength{\abovedisplayskip}{3pt}
\setlength{\belowdisplayskip}{2pt}
\begin{equation}
\mathcal{L}_{\text{CoT-SFT}} = -\mathbb{E}_{(q, \text{CoT}, A) \sim \mathcal{D}} \left[ \log p_\theta(\text{CoT}, A \mid q, \mathcal{O}) \right]
\end{equation}
}

where $q$ is the task instruction, $\text{CoT}$ is the reasoning trace, $A$ is the final answer, and $\mathcal{O}$ represents scene object features.
This objective teaches the model to internalize systematic decomposition patterns while maintaining transparency through explicit intermediate steps.

\subsubsection{Open-Set and Complex Reasoning Generalization}
\label{sec:cot-rl}

Finally, we employ Chain-of-Thought Reinforcement Learning (CoT-RL) to extend Stage 2's systematic reasoning to handle real-world scenarios
(e.g., ScanRefer~\cite{chen2020scanrefer} and Multi3DRefer~\cite{zhang2023multi3drefer})
involving open-vocabulary concepts (e.g., "comfortable") and deep nesting where intermediate supervision is absent.
For example, "find the comfortable furniture on the kitchen counter and besides the white fridge" requires understanding open-set attributes ("comfortable") and complex spatial reasoning beyond programs with 2 nesting level.
Our CoT-RL stage addresses this challenge via Group Relative Policy Optimization (GRPO)~\cite{shao2024deepseekmathpushinglimitsmathematical} with a composite reward function tailored for spatial grounding.

First, to overcome the sparsity of IoU metrics which provide no guidance for disjoint predictions, we introduce a \textbf{Soft Grounding Reward} that evaluates configuration similarity:
{
\setlength{\abovedisplayskip}{3pt}
\setlength{\belowdisplayskip}{3pt}
\begin{equation}
    R_\text{grounding} = \underbrace{e^{-\alpha \| \bm{x}_{\text{pred}} - \bm{x}_{\text{gt}} \|_2}}_{\text{location similarity}} + \underbrace{e^{-\alpha \left\|\frac{\bm{s}_{\text{pred}} - \bm{s}_{\text{gt}}}{\bm{s}_{\text{gt}}}\right\|_1}}_{\text{size similarity}}
\end{equation}
}
where $\bm{x}_{\text{pred}}, \bm{x}_{\text{gt}} \in \mathbb{R}^3$ are predicted and ground-truth object centers, $\bm{s}_{\text{pred}}, \bm{s}_{\text{gt}} \in \mathbb{R}^3$ are object sizes, and $\alpha=2$ controls decay rate.
This offers dense feedback based on spatial proximity, enabling effective exploration even when initial predictions yield zero overlap.
Concurrently, to prevent the policy from collapsing into direct answering and losing interpretability, we incorporate a binary \textbf{Format Reward}: $R_\text{format}(o) = 1$ if the model response contains valid \textit{plan} and \textit{thinking tags} while being not too short, and $0$ otherwise. This ensures the model maintains systematic reasoning structure and avoids thinking degradation.

\noindent \textbf{Policy Generation and Update.}
We optimize the policy using GRPO. For each instruction $q$, we sample $N$ responses $\{o_1, \ldots, o_N\}$ from the current policy $\pi_\theta$ and compute their rewards $r_i = R_\text{format}(o_i) + R_\text{grounding}(o_i)$.
We then compute group-normalized advantages
$A_i = \frac{r_i-\operatorname{mean}(\{r_1,...,r_N\})}{\operatorname{std}({\{r_1,...,r_N\}}}$.
The policy $\pi_\theta$ is updated via clipped surrogate objective:
{
\setlength{\abovedisplayskip}{3pt}
\setlength{\belowdisplayskip}{2pt}
\begin{align}
\mathcal{L}_{\text{GRPO}}(\theta) &= \mathbb{E}_{q,\{o_i\}} \Bigg[ \frac{1}{N} \sum_{i=1}^{N} \bigg( \min \Big( \frac{\pi_\theta(o_i|q)}{\pi_{\theta_{\text{old}}}(o_i|q)} A_i, \notag \\
&\qquad \text{clip} \Big( \frac{\pi_\theta(o_i|q)}{\pi_{\theta_{\text{old}}}(o_i|q)}, 1-\varepsilon, 1+\varepsilon \Big) A_i \Big) \bigg) \notag \\
&\qquad - \beta \mathcal{D}_{\text{KL}}(\pi_\theta \| \pi_{\text{ref}}) \Bigg]
\end{align}
}

\subsection{Modular Inference Enhancement via Symbolic Decoupling}
\label{sec:modularity}

A unique advantage of \methodName{} unattainable by black-box 3D MLLMs is the explicit decoupling of planning (reasoning) and execution (perception and spatial verification).
While standard 3D MLLMs fuse these processes into an opaque latent space, our neuro-symbolic architecture maintains a transparent interface, enabling plug-and-play integration of external state-of-the-art components during inference without retraining.

\noindent \textbf{External Plan Injection.}
Our structural CoT acts as a universal control interface that allows self-generated plans to be seamlessly replaced by external planners.
During inference, we can inject refined decomposition plans from frontier LLMs (e.g., GPT-4~\cite{openai2023gpt4}) into the reasoning trace.
This enables our model to function as a \textit{visual execution engine}, leveraging the superior planning capabilities of larger models to solve complex queries that exceed its internal capacity.

\noindent \textbf{Perception Module Replacement.}
Similarly, our framework supports modular upgrades of execution components.
As an example, we demonstrate replacing the \lstinline|scene()| primitive execution which acts as a universal interface for perception models.
We replace the default \lstinline|scene()| exeuction by model itself with outputs from state-of-the art instance segmentation models (e.g., SegDINO3D~\cite{qu2025segdino3d3dinstancesegmentation}) by simply formatting their outputs to match the CoT trace.
This design facilitates continuous improvement of our reasoning system from advancements in 3D perception.

\section{Experiments}
\label{sec:experiments}

\subsection{Evaluation and Implementation Details}
\label{subsec:setup}

\noindent\textbf{Datasets and Metrics.}
Following previous NS methods \cite{Hsu2023NS3DNG, Feng2024NaturallyS3}, we evaluate \methodName{} on spatial reasoning 3D vision-language benchmarks, namely ScanRefer~\cite{chen2020scanrefer} and Multi3DRefer~\cite{zhang2023multi3drefer}.
For ScanRefer, we report accuracy at IoU thresholds of 0.25 and 0.5 (Acc@0.25,
Acc@0.5), and F1 scores for Multi3DRefer.
To further assess generalizability, we also evaluate our method on SQA3D~\cite{ma2022sqa3d} for situated question answering (Exact Match) and Scan2Cap~\cite{chen2021scan2cap} for dense captioning (CiDEr@IoU scores).

\noindent\textbf{Implementation Details.}
We implement \methodName{} on a standard 8B MLLM backbone~\cite{yang2025qwen3technicalreport},
employing AdamW~\cite{loshchilov2017decoupled} and Muon~\cite{jordan2024muon, liu2025muonscalablellmtraining} optimizer with LoRA~\cite{hu2022lora} for efficient fine-tuning.
Please refer to the supplementary for full details.

\begin{table*}[t]
    \centering
    \fontsize{7.5}{10}\selectfont
    \caption{\textbf{3D Spatial Reasoning Performance Comparison.} We compare methods based on their output format: `Head' denotes methods using an extra grounding decoder, while `Text' denotes methods that output box boundaries or identifiers directly as plain text. Best and second best are in \textbf{bold} and \underline{underlined}. $^\dagger$: with modular inference enhancement.
    }
    \label{tab:vg_results}
    \begin{tabularx}{\linewidth}{l c Y Y Y Y}
        \toprule
        \multirow{2}{*}{\textbf{Method}} & \multirow{2}{*}{\textbf{Output}} & \multicolumn{2}{c}{\textbf{ScanRefer}} & \multicolumn{2}{c}{\textbf{Multi3DRefer}} \\
        \cmidrule(lr){3-4} \cmidrule(lr){5-6}
        & & Acc@0.25 & Acc@0.5 & F1@0.25 & F1@0.5 \\
        \midrule
        \multicolumn{6}{l}{\textit{Neuro-Symbolic Methods}} \\
        NS3D~\cite{Hsu2023NS3DNG} & Head & 22.4 & - & - & - \\
        LARC~\cite{Feng2024NaturallyS3} & Head & 32.9 & - & - & - \\
        LaSP~\cite{mi-etal-2025-lasp}  & Text & 49.2 & - & - & - \\
        \midrule
        \multicolumn{6}{l}{\textit{Specialist Methods}} \\
        ScanRefer~\cite{chen2020scanrefer} & Head & 37.3 & 24.3 & - & - \\
        M3DRef-CLIP~\cite{zhang2023multi3drefer} & Head & 51.9 & 44.7 & 42.8 & - \\
        3D-VisTA~\cite{zhu20233dvista} & Head & 50.6 & 45.8 & - & - \\
        SceneVerse~\cite{jia2024sceneverse} & Head & - & 48.1 & - & - \\
        \midrule
        \multicolumn{6}{l}{\textit{3D MLLMs}} \\
        Grounded 3D-LLM~\cite{chen2024grounded3dllmreferenttokens} & Head & 48.6 & 44.0 & 44.7 & 40.8 \\
        LLaVA-3D~\cite{zhu2025llava3dsimpleeffectivepathway} & Head & 50.1 & 42.7 & 49.8 & 43.6 \\
        PQ3D~\cite{zhu2024pq3d} & Head & 57.0 & 51.2 & - & 50.1 \\
        Video-3D LLM~\cite{zheng2024video3dllmlearningpositionaware} & Head & 58.1 & \underline{51.7} & 58.0 & 52.7 \\
        3D-LLaVA~\cite{Deng_2025_CVPR} & Text & 51.2 & 40.6 & - & - \\
        Chat-Scene~\cite{huang2024chat} & Text & 55.5 & 50.2 & 57.1 & 52.4 \\
        Inst3D-LMM~\cite{Inst3D-LMM} & Text & 57.8 & 51.6 & 58.3 & 53.5 \\
        \rowcolor{LightBlue}
        \methodName{} (Ours) & Text & \underline{58.4} & 51.2 & \underline{59.2} & \underline{53.8} \\
        \rowcolor{LightBlue}
        \methodName{}$^\dagger$ (Ours) & Text & \textbf{60.5} & \textbf{53.2} & \textbf{60.9} & \textbf{55.2} \\
        \bottomrule
    \end{tabularx}
    \label{tab:grounding_main}
\end{table*}

\subsection{Performance on 3D Spatial Reasoning}
\label{subsec:grounding}

As illustrated in \Cref{tab:grounding_main}, compared to state-of-the-art \textbf{3D MLLMs} (e.g., Chat-Scene, Inst3D-LMM), \methodName{} achieves superior performance, notably surpassing the strongest baseline on both ScanRefer and Multi3DRefer datasets. This gain empirically demonstrate the effectiveness of our distilled systematic reasoning, which enables more precise spatial verification than the implicit "black-box" logic of standard MLLMs.
Conversely, traditional \textbf{Neuro-Symbolic methods} (e.g., NS3D) struggle severely on natural language datasets like ScanRefer and Nr3D due to their closed-set vocabulary constraints and rigid program parsers. It fails to generalize to the linguistic diversity of real-world instructions, resulting in subpar performance.
These results confirms that \methodName{} could combine the best of both worlds: the systematic reasoning of symbolic programs and the semantic flexibility of LLMs.

\subsection{Generalization to Open-Set Concepts}
\label{subsubsec:openset}

Traditional NS methods~\cite{Hsu2023NS3DNG,Feng2024NaturallyS3} are fundamentally limited by their reliance on closed-set vocabularies. Does \methodName{}'s distilled reasoning capabilities generalize to open-set concepts unseen during training?
To evaluate this, we leverage the pairing of Sr3D (with synthetic instructions) and Nr3D (with natural human language instructions) datasets~\cite{achlioptas2020referit_3d}, which share identical scenes but differ in linguistic complexity.
We train our 3D MLLM \textit{exclusively on the Sr3D} (Stage 2-only) and evaluate it in a \textbf{training-free zero-shot} setting on the Nr3D with unseen concepts and instruction structures.
This setup strictly tests whether the reasoning patterns distilled from programs can generalize to free-form descriptions that could defy symbolic parsing.

As shown in Table~\ref{tab:openset_comparison}, \methodName{} achieves \textbf{36.5\%} accuracy on Nr3D without seeing any real-world annotations.
Remarkably, this zero-shot performance surpasses the \textbf{fully supervised} NS3D baseline (trained on Nr3D) by \textbf{+2.6\%}.
While NS3D fails to learn and generalize open-vocabulary concepts (e.g., "messy", "cozy"), \methodName{} leverages the LLM's semantic knowledge to interpret these nuanced descriptors.
This empirically validates that our paradigm could break the "vocabulary bottleneck" itraditional neuro-symbolic systems, enabling effective sim-to-real transfer of the reasoning capabilities.

\begin{table}[t]
\fontsize{7.5}{10}\selectfont
    \centering
    \caption{\textbf{Zero-shot Generalization on Open-Set Concepts.} We compare \methodName{} (Zero-shot, never seen Nr3D) against NS methods. \showbest}
    \label{tab:openset_comparison}
    \begin{tabularx}{\linewidth}{l Y c Y}
        \toprule
        \textbf{Method} & \textbf{Training Data} & \textbf{Setting} & \textbf{Acc} \\
        \midrule
        NS3D~\cite{Hsu2023NS3DNG} & Sr3D & Zero-shot & 7.3 \\
        NS3D~\cite{Hsu2023NS3DNG} & Nr3D & Supervised & 33.9 \\
        R2G~\cite{Li2025r2g} & Sr3D & Zero-shot & 25.8 \\
        \midrule
        \textbf{\methodName{} (Ours)} & Sr3D & Zero-shot & \textbf{36.5} \\
        \bottomrule
    \end{tabularx}
\end{table}

\begin{table}[t]
\centering
\fontsize{7.5}{10}\selectfont
\caption{Results on Scan2Cap (dense captioning) and SQA3D (situated question answering).  Best and second best are in \textbf{bold} and \underline{underlined}.}
\label{tab:cross_task}
\begin{tabularx}{\linewidth}{l YYYY}
\toprule
\multirow{3}{*}{\textbf{Method}}
  & \multicolumn{3}{c}{\textbf{Scan2Cap}}
  & \textbf{SQA} \\
\cmidrule(lr){2-4} \cmidrule(lr){5-5}
  & \multicolumn{2}{c}{ScanRefer}
  & Nr3D
  & \\
\cmidrule(lr){2-3} \cmidrule(lr){4-4}
  & C@.25 & C@.5 & C@.5 & EM \\
\midrule
\multicolumn{5}{l}{\textit{Specialist Methods}} \\
Scan2Cap~\cite{chen2021scan2cap}
  & 56.8 & 39.1 & 27.5 & -- \\
SQA3D~\cite{ma2022sqa3d}
  & -- & -- & -- & 47.2 \\
3D-VisTA~\cite{zhu20233dvista}
  & -- & 61.6 & -- & 48.5 \\
V2C-DETR~\cite{chen2023end}
  & 71.4 & 61.8 & 43.8 & -- \\
V2C-DETR++~\cite{chen2023vote2capdetr}
  & 76.4 & 67.6 & 47.1 & -- \\
BridgeQA~\cite{Mo_Liu_2024}
  & -- & -- & -- & 52.9 \\
PQ3D~\cite{zhu2024pq3d}
  & -- & 80.3 & -- & 47.1 \\
\midrule
\multicolumn{5}{l}{\textit{3D MLLMs}} \\
LL3DA~\cite{chen2023ll3da}
  & 74.2 & 65.2 & 51.2 & -- \\
LEO~\cite{huang2023embodied}
  & -- & 72.4 & -- & 50.0 \\
SceneLLM~\cite{fu2024scenellm}
  & -- & -- & -- & 53.6 \\
ChatScene~\cite{huang2024chat}
  & -- & 77.2 & -- & 54.6 \\
LEGO~\cite{mo2025lego}
  & \underline{84.7} & 78.6 & \underline{61.4} & 54.8 \\
LLaVA-3D~\cite{zhu2025llava3dsimpleeffectivepathway}
  & -- & 79.2 & -- & \underline{55.6} \\
Inst3D-LMM~\cite{Inst3D-LMM}
  & -- & 79.7 & -- & -- \\
Video-3D LLM~\cite{zheng2024video3dllmlearningpositionaware}
  & -- & \underline{83.8} & -- & \textbf{58.6} \\
\cellcolor{LightBlue}\textbf{\methodName{} (Ours)}
  & \cellcolor{LightBlue}\textbf{90.6}
  & \cellcolor{LightBlue}\textbf{84.1}
  & \cellcolor{LightBlue}\textbf{68.1}
  & \cellcolor{LightBlue}\textbf{58.6} \\
\bottomrule
\end{tabularx}
\vspace{1em}
\end{table}
\begin{table}[t]
\centering
\fontsize{7.5}{10}\selectfont
\caption{\textbf{Design Ablations.} We evaluate the impact of our curriculum stages, RL reward components and whether to think, on ScanRefer and Multi3DRefer datasets. \showbest}
\label{tab:ablation_design}
\begin{tabularx}{\linewidth}{l Y Y}
\toprule
\multirow{2}{*}{\textbf{Variant}} & \textbf{ScanRefer} & \textbf{Multi3DRefer} \\
& Acc@0.25 & F1@0.25 \\
\midrule
\rowcolor{LightBlue} \textbf{\textbf{\methodName{}}} & \textbf{58.4} & \textbf{59.2} \\
\midrule
\multicolumn{3}{l}{\textit{Training Curriculum Stages}} \\
w/o Stage 3 (No CoT-RL) & 51.5 & 55.3 \\
~~\textit{\scriptsize{(Stage 3 $\to$ Direct SFT)}} & \textcolor{BrickRed}{-6.9} & \textcolor{BrickRed}{-3.9} \\
w/o Stage 2 (No CoT-SFT) & 48.2 & 36.7 \\
~~\textit{\scriptsize{(Direct Alignment $\to$ CoT-RL)}} & \textcolor{BrickRed}{-10.2} & \textcolor{BrickRed}{-22.5} \\
\midrule
\multicolumn{3}{l}{\textit{RL Reward Components}} \\
w/o Format Reward & 55.7 & 57.1 \\
~~\textit{\scriptsize{(Grounding Reward Only)}} & \textcolor{BrickRed}{-2.7} & \textcolor{BrickRed}{-2.1} \\
w/o Soft Grounding Reward & 57.7 & 58.7 \\
~~\textit{\scriptsize{(Use IoU Reward Instead)}} & \textcolor{BrickRed}{-0.7} & \textcolor{BrickRed}{-0.5} \\
\midrule
\multicolumn{3}{l}{\textit{Inference Strategy}} \\
w/o Thinking & 56.8 & 58.2 \\
~~\textit{\scriptsize{(Direct Answer Output)}} & \textcolor{BrickRed}{-1.6} & \textcolor{BrickRed}{-1.0} \\
\bottomrule
\end{tabularx}
\end{table}


\begin{table}[t]
\centering
\fontsize{7.5}{10}\selectfont
\caption{\textbf{RL Benefit by Reasoning Complexity} on ScanRefer (Acc@0.5).
  CoT-RL yields larger gains on longer, more complex reasoning chains
  where Stage~2 CoT-SFT lacks supervised traces.}
\label{tab:rl_complexity}
\begin{tabularx}{\linewidth}{l YYYY}
\toprule
\textbf{Complexity} & \textbf{Steps} & \textbf{SFT-only} & \textbf{CoT-RL} & \textbf{$\Delta$} \\
\midrule
Short  & $\leq$4 & 47.2 & 45.4 & \textcolor{BrickRed}{$-$1.8} \\
Medium & $=$5    & 50.7 & 52.2 & \textcolor{ForestGreen}{+1.5} \\
Long   & $\geq$6 & 45.2 & 47.9 & \textcolor{ForestGreen}{+2.7} \\
\bottomrule
\end{tabularx}
\vspace{1em}
\end{table}

\begin{table}[t]
\centering
\fontsize{7.5}{10}\selectfont
\caption{\textbf{Modularity Analysis.}
We demonstrate the modularity of \methodName{} by replacing its planning and perception executions at inference time. Best non-oracle results are in \textbf{bold}. $^\star$: M3DRef stands for Multi3DRefer. $^\dagger$: with modular inference enhancement.
}
\label{tab:modularity}
\begin{tabularx}{\linewidth}{l l Y c}
\toprule
\multirow{2}{*}{\textbf{Module}} & \multirow{2}{*}{\textbf{Source}} & \textbf{ScanRefer} & \textbf{M3DRef$^\star$} \\
& & Acc@0.25 & F1@0.25 \\
\midrule
\multicolumn{2}{l}{\cellcolor{LightBlue}\textbf{\methodName{}}} & \cellcolor{LightBlue}58.4 & \cellcolor{LightBlue}59.2 \\
\multicolumn{2}{l}{~~\scriptsize{\textit{(Self-Plan + Self-Perception)}}} & & \\
\midrule
\multirow{2}{*}{\textbf{Planning}}
 & External Planner  & 58.6 & 59.5 \\
 & ~~\scriptsize{\textit{(Self-Plan \(\to\) Claude 4.5 Opus})}& \textcolor{ForestGreen}{+\textbf{0.2}} & \textcolor{ForestGreen}{+\textbf{0.3}} \\
\midrule
\multirow{6}{*}{\textbf{Perception}}
 & Mask3D~\cite{Schult23ICRA} & 58.5 & 58.3 \\
 & ~~\scriptsize{\textit{(External Perception)}} & \textcolor{ForestGreen}{+\textbf{0.1}} & \textcolor{BrickRed}{-\textbf{0.9}} \\
 & SegDINO3D~\cite{qu2025segdino3d3dinstancesegmentation} & 60.4 & 60.6 \\
 & ~~\scriptsize{\textit{(Strong External Perception)}} & \textcolor{ForestGreen}{+\textbf{2.0}} & \textcolor{ForestGreen}{+\textbf{1.4}} \\
 & \cellcolor{gray!15}Oracle (Ground-Truth Labels) & \cellcolor{gray!15}61.3 & \cellcolor{gray!15}61.3 \\
 & \cellcolor{gray!15}~~\scriptsize{\textit{(Upper Bound)}} & \cellcolor{gray!15}\textcolor{ForestGreen}{+\textbf{2.9}} & \cellcolor{gray!15}\textcolor{ForestGreen}{+\textbf{2.1}} \\
 \midrule
\multicolumn{2}{l}{\cellcolor{LightBlue}\textbf{\methodName{}$^\dagger$}} & \cellcolor{LightBlue}\textbf{60.5} & \cellcolor{LightBlue}\textbf{60.9} \\
\multicolumn{2}{l}{~~\scriptsize{\textit{(Full Modular Enhancement)}}} & \textcolor{ForestGreen}{+\textbf{2.1}} & \textcolor{ForestGreen}{+\textbf{1.7}} \\
\bottomrule
\end{tabularx}
\vspace{1em}
\end{table}

\subsection{Generalization to More Tasks}
\label{sec:cross_task}

To validate that our curriculums is a general training paradigm rather than a grounding-specific pipeline, we further evaluate \methodName{} on two additional 3D scene understanding tasks (i.e., SQA3D~\cite{ma2022sqa3d} for \textbf{Situated Question Answering} and Scan2Cap~\cite{chen2021scan2cap} for \textbf{Dense Captioning}) by simply swapping the outcome reward in Stage~3: exact-match (EM) for question answering and CIDEr for dense captioning.
No modification to the symbolic reasoning pipeline in Stages~1 and 2 is required.
Specifically, SQA3D requires 3D MLLM to conduct perspective-dependent spatial reasoning, answering questions based on the asker's position, and is evaluated by answer exact-match (EM) score.
Scan2Cap requires generating spatially grounded natural language descriptions for objects in 3D scenes, evaluated by CiDEr scores conditioned on localization accuracy.

As shown in \Cref{tab:cross_task}, \methodName{} matches or surpasses prior best methods across all metrics on both tasks, despite being primarily designed for spatial grounding, confirming that our three-stage curriculum is a training paradigm generalizable to various 3D scene understanding tasks.

\subsection{Ablation Studies}
\label{subsec:ablation}

We conduct comprehensive ablation studies on ScanRefer and Multi3DRefer to validate our curriculum design, reward formulation, and modular architecture.

\noindent \textbf{Impact of Curriculum Stages.}
We first investigate the necessity of our progressive training pipeline (\Cref{tab:ablation_design}).
\textbf{(1) Efficacy of RL (Stage 3):} Replacing the CoT-RL stage with direct-answer SFT on the same downstream data results in a significant performance drop of 6.9\% and 3.9\% on ScanRefer and Multi3DRefer, respectively. This confirms that while Stage 2 teaches the "syntax" of reasoning, RL is crucial for adapting these patterns to the complex linguistic structures of real-world instructions.
\textbf{(2) Necessity of Reasoning Injection (Stage 2):} Attempting to train CoT-RL directly from the alignment stage leads to a catastrophic performance degradation. This might due to VLM's lack of inherent knowledge of 3D spatial verification, and without the "warm start" provided by structured reasoning in symbolic programs, the RL agent struggles to explore the reasoning space from scratch effectively.

\noindent \textbf{Analysis of RL Rewards.}
We further dissect the contribution of our reward components with following observations (\Cref{tab:ablation_design}):
\textbf{(1) Format Reward is Critical:} Removing the format constraint causes moderate performance drop with response length degration. Qualitative inspections reveals that without this constraint, the model frequently suffers from "structure collapse" that either only outputting instructions again or skipping the reasoning process entirely, since they are the "shortcut" path to maximize reward at the early RL stage.
\textbf{(2) Soft Grounding Reward Benefits:} Replacing our continuous configuration-based reward with sparse IoU feedback results in a slight but consistent 0.7\% and 0.5\% decline. While not the most effective factor, dense feedback helps guide RL optimization.

\noindent \textbf{Inference Strategy Analysis.}
Does the model actually need to "think"? When we force the fully trained model to output the answer directly without generating the thinking trace, performance drops by \textbf{1.6\%} and \textbf{1.0\%} (\Cref{tab:ablation_design}). Interestingly, this "direct" inference still significantly outperforms the Stage 2 (SFT) baseline (56.8 vs 51.5 and 58.2 vs. 55.3), suggesting that RL has improved the model's internal representations, but the explicit thinking procedure remains essential for unlocking its full spatial reasoning potential.

\noindent \textbf{Impact of Reasoning Complexity on RL Benefit.}
To understand \emph{where} Stage~3 (CoT-RL Stage) contributes most, we group ScanRefer validation samples by reasoning complexity, estimated by the number of plan steps in the generated CoT, and compare models trained with and without RL.
As shown in~\Cref{tab:rl_complexity}, a clear trend emerges: \textbf{RL benefit increases with reasoning complexity.}
For simple queries ($\leq$4 steps), SFT from synthetic programs already captures sufficient patterns, and RL exploration introduces marginal noise. For complex queries requiring multi-step spatial verification ($\geq$6 steps), CoT-RL provides consistent improvement (+2.7\%). This is because CoT-SFT lacks supervised traces for longer programs, while Stage~3's outcome-based RL can reward correct longer traces discovered through exploration, effectively providing supervision signals for complex reasoning chains that CoT-SFT cannot cover. This further validates our Stage~3's role: extending systematic neuro-symbolic reasoning to complex, real-world instructions where dense symbolic supervision is unavailable.

\noindent \textbf{Modularity and Scalability.}
Finally, we analyze the source of our model's performance via module replacement (\Cref{tab:modularity}).
\textbf{(1) Planning Saturation:} Replacing self-generated plans with those from a powerful external LLM (Claude 4.5 Opus~\cite{opus4_5}) yields only marginal gains. This indicates that our distilled planner is already highly robust, effectively leveraging the inherent instruction parsing capabilities of the base LLM.
\textbf{(2) Perception Scalability:} In contrast, replacing the perception module with a stronger semantic recognition model (SegDINO3D~\cite{qu2025segdino3d3dinstancesegmentation}) significantly boosts performances, nearly matching the Oracle upper bound.
This demonstrates that visual perception accuracy limit our current system more than planning failures.
However, this limitation effectively highlights the \textit{scalability} of our modular architecture: unlike black-box end-to-end models, \methodName{} allows us to address this visual bottleneck by seamlessly integrating stronger 3D perception models as they emerge without re-training.

\subsection{Qualitatives of Emergent Reasoning Behaviors}
\label{subsec:qualitative}

\begin{figure}[!ht]
\centering
\begin{promptbox}
\small 
\think{[think]}

I need to find the object described as: "this \textbf{\textcolor{orange}{beige chair}} is next to the coat rack. it is also to the left of the table and lamp."

\commenttt{// Planning}

Let's plan my next steps: Find objects of category \textbf{\textcolor{orange}{'beige chair'}}, 'coat rack', 'table', and 'lamp'; Check which \textbf{\textcolor{orange}{beige chair(s) are 'next to' the coat rack(s) and 'to the left of' the table(s) and lamp(s)}}.

\commenttt{// Execute scene()}

I see 33 object(s) in the scene: 0 (couch), ... \textcolor{gray}{[details omitted]}

\commenttt{// Execute filter(beige chair)}

Looking for 'beige chair'.
I find 2 object(s): 3, 5, at ...

\commenttt{// Execute filter(coat rack)}

Looking for 'coat rack’.
I find 1 object(s): 17 at...

\commenttt{// Execute filter(table)}

Looking for 'table’.
I find 6 object(s): 7, 8, 9, ...

\commenttt{// Execute filter(lamp)}

Looking for 'lamp’.
I find 3 object(s): 16, 25, 31, at ...

\commenttt{// Execute relate(filter(beige chair), union(filter(table), filter(lamp)), next to)}

Analyzing... Object 5 is next to Object 17 and to the left of Object 14 and 16.
After analysis, I find 1 object(s) 'next to' ...

\commenttt{// Execute relate(filter(beige chair), filter(coat rack), left)}

After analysis, I find 1 object(s) 'to the left of'...

\commenttt{// Execute intersection(relate(...), relate(...))}

Now, I'll check which beige chair(s) are 'next to' the coat rack(s) and 'to the left of' the table(s) and lamp(s): \textbf{\textcolor{orange}{Object 5 (chair)}} matches both relations.

\answer{[answer]}

I've located 1 object(s) as described:
Object 5: Coordinates (3.47, 3.18, 2.49), dimensions 0.80 x 0.51 x 1.47
\end{promptbox}
\caption{Example CoT revealing emerging reasoning patterns in CoT-RL stage. Object location details are omitted for brevity.}
\label{fig:cot_emerging}
\end{figure}

A key hypothesis of our curriculum is that CoT-RL (Stage 3) enables generalization beyond the fixed templates of Stage 2.
As shown in \Cref{fig:cot_emerging}, qualitative analysis reveals that \methodName{} spontaneously \textbf{invents new reasoning patterns} to bridge the gap between symbolic rigor and linguistic flexibility.
First, it breaks the "vocabulary bottleneck" by adapting the \texttt{filter} primitive to handle \textbf{open-vocabulary descriptors} (e.g., generating \texttt{filter(beige chair)} instead of rigid categories), effectively leveraging the VLM's semantic knowledge.
Second, the model successfully constructs plans for more \textbf{deeply nested queries} unseen during SFT. To resolve multi-condition constraints (e.g., "next to A \textbf{AND} left of B"), it spontaneously emerges new logical primitives such as \texttt{intersection} and \texttt{union}, which are never explicitly taught in the curriculum.
This indicates that \methodName{} has not merely memorized program templates but has internalized the underlying "syntax" of spatial logic, allowing it to dynamically compose novel primitives to match the complexity of real-world instructions.

\section{Conclusion}
\label{sec:concl}

We presented \methodName{}, a neuro-symbolic 3D MLLM that bridges the gap between interpretable but closed-set symbolic methods and flexible but opaque end-to-end MLLMs.
Our key insight is that reasoning patterns can be distilled from symbolic programs into natural language chain-of-thought and integrated into MLLM, enabling systematic spatial reasoning with open-vocabulary flexibility.
Through a three-stage curriculum progressing from perception alignment and symbolic reasoning injection to RL-based generalization, \methodName{} learns to decompose queries, verify spatial relations step-by-step, and adapt to complex real-world instructions.
Experiments demonstrate that \methodName{} surpasses prior neuro-symbolic methods and matches state-of-the-art 3D MLLMs, while preserving key virtues: transparent reasoning traces for interpretability and modular architecture for scalable upgrades.
We hope this work offers a initial stepstone toward interpretable and useful embodied agents.

\section*{Impact Statement}
\label{sec:impact}

This work advances interpretable 3D spatial reasoning through neuro-symbolic Chain-of-Thought distillation. Our framework seeks to establish a more systematic and verifiable grounding of objects within 3D environments, offering potential societal benefits for assistive robotics, indoor navigation, and embodied AI systems, where transparent failure analysis is critical for trust calibration, safety, and effective debugging.

\section*{Acknowledgment}
\label{sec:ack}
This work was supported by the grants from the Beijing
Natural Science Foundation 4252040, Beijing Nova Program and National Natural Science Foundation of China 62372014 and CAAI-Tencent Rhino-Bird Open Research Fund.

\bibliography{main}

\begin{thebibliography}{56}
\providecommand{\natexlab}[1]{#1}
\providecommand{\url}[1]{\texttt{#1}}

\bibitem[Achlioptas et~al.(2020)Achlioptas, Abdelreheem, Xia, Elhoseiny, and
  Guibas]{achlioptas2020referit_3d}
Achlioptas, P., Abdelreheem, A., Xia, F., Elhoseiny, M., and Guibas, L.
\newblock Referit3d: Neural listeners for fine-grained 3d object identification
  in real-world scenes.
\newblock \emph{16th European Conference on Computer Vision (ECCV)}, 2020.

\bibitem[Anthropic(2025)]{opus4_5}
Anthropic.
\newblock Introducing claude opus 4.5.
\newblock \url{https://www.anthropic.com/news/claude-opus-4-5}, 2025.

\bibitem[Chen et~al.(2020)Chen, Chang, and Nie{\ss}ner]{chen2020scanrefer}
Chen, D.~Z., Chang, A.~X., and Nie{\ss}ner, M.
\newblock Scanrefer: 3d object localization in rgb-d scans using natural
  language.
\newblock \emph{16th European Conference on Computer Vision (ECCV)}, 2020.

\bibitem[Chen et~al.(2023)Chen, Zhu, Chen, Lei, Yu, and Chen]{chen2023end}
Chen, S., Zhu, H., Chen, X., Lei, Y., Yu, G., and Chen, T.
\newblock End-to-end 3d dense captioning with vote2cap-detr.
\newblock In \emph{Proceedings of the IEEE/CVF Conference on Computer Vision
  and Pattern Recognition}, pp.\  11124--11133, 2023.

\bibitem[Chen et~al.(2024{\natexlab{a}})Chen, Chen, Zhang, Li, Yu, Fei, Zhu,
  Fan, and Chen]{chen2023ll3da}
Chen, S., Chen, X., Zhang, C., Li, M., Yu, G., Fei, H., Zhu, H., Fan, J., and
  Chen, T.
\newblock Ll3da: Visual interactive instruction tuning for omni-3d
  understanding reasoning and planning.
\newblock In \emph{Proceedings of the IEEE/CVF Conference on Computer Vision
  and Pattern Recognition (CVPR)}, pp.\  26428--26438, 2024{\natexlab{a}}.

\bibitem[Chen et~al.(2024{\natexlab{b}})Chen, Zhu, Li, Chen, Guo, Lei, Yu, Li,
  and Chen]{chen2023vote2capdetr}
Chen, S., Zhu, H., Li, M., Chen, X., Guo, P., Lei, Y., Yu, G., Li, T., and
  Chen, T.
\newblock Vote2cap-detr++: Decoupling localization and describing for
  end-to-end 3d dense captioning.
\newblock \emph{IEEE Transactions on Pattern Analysis and Machine
  Intelligence}, 46\penalty0 (11):\penalty0 7331--7347, 2024{\natexlab{b}}.

\bibitem[Chen et~al.(2024{\natexlab{c}})Chen, Yang, Huang, Wang, Xu, Lyu, Lin,
  and Pang]{chen2024grounded3dllmreferenttokens}
Chen, Y., Yang, S., Huang, H., Wang, T., Xu, R., Lyu, R., Lin, D., and Pang, J.
\newblock Grounded 3d-llm with referent tokens, \emph{arXiv preprint
  arXiv:2405.10370}, 2024{\natexlab{c}}.

\bibitem[Chen et~al.(2021)Chen, Gholami, Nie{\ss}ner, and
  Chang]{chen2021scan2cap}
Chen, Z., Gholami, A., Nie{\ss}ner, M., and Chang, A.~X.
\newblock Scan2cap: Context-aware dense captioning in rgb-d scans.
\newblock In \emph{Proceedings of the IEEE/CVF Conference on Computer Vision
  and Pattern Recognition}, pp.\  3193--3203, 2021.

\bibitem[Dai et~al.(2017)Dai, Chang, Savva, Halber, Funkhouser, and
  Nie{\ss}ner]{dai2017scannet}
Dai, A., Chang, A.~X., Savva, M., Halber, M., Funkhouser, T., and Nie{\ss}ner,
  M.
\newblock Scannet: Richly-annotated 3d reconstructions of indoor scenes.
\newblock In \emph{Proceedings of the IEEE conference on computer vision and
  pattern recognition}, pp.\  5828--5839, 2017.

\bibitem[Deng et~al.(2025)Deng, He, Jiang, Wang, Dayoub, and
  Reid]{Deng_2025_CVPR}
Deng, J., He, T., Jiang, L., Wang, T., Dayoub, F., and Reid, I.
\newblock 3d-llava: Towards generalist 3d lmms with omni superpoint
  transformer.
\newblock In \emph{Proceedings of the IEEE/CVF Conference on Computer Vision
  and Pattern Recognition (CVPR)}, pp.\  3772--3782, 2025.

\bibitem[Feng et~al.(2024)Feng, Hsu, Liu, and Wu]{Feng2024NaturallyS3}
Feng, C., Hsu, J., Liu, W., and Wu, J.
\newblock Naturally supervised 3d visual grounding with language-regularized
  concept learners.
\newblock \emph{2024 IEEE/CVF Conference on Computer Vision and Pattern
  Recognition (CVPR)}, pp.\  13269--13278, 2024.

\bibitem[Fu et~al.(2025)Fu, Liu, Chen, Nie, and Xiong]{fu2024scenellm}
Fu, R., Liu, J., Chen, X., Nie, Y., and Xiong, W.
\newblock Scene-llm: Extending language model for 3d visual reasoning.
\newblock In \emph{Proceedings of the Winter Conference on Applications of
  Computer Vision (WACV)}, pp.\  2195--2206, 2025.

\bibitem[Gong et~al.(2026)Gong, Gao, Song, Sun, Zhou, and Li]{Gong2026}
Gong, T., Gao, S., Song, Q., Sun, Q., Zhou, H., and Li, J.
\newblock Towards reliable multimodal intelligence via uncertainty-aware
  inference.
\newblock \emph{Chinese Journal of Electronics}, 35\penalty0 (4):\penalty0
  1--17, 2026.

\bibitem[Hsu et~al.(2023)Hsu, Mao, and Wu]{Hsu2023NS3DNG}
Hsu, J., Mao, J., and Wu, J.
\newblock Ns3d: Neuro-symbolic grounding of 3d objects and relations.
\newblock \emph{2023 IEEE/CVF Conference on Computer Vision and Pattern
  Recognition (CVPR)}, pp.\  2614--2623, 2023.

\bibitem[Hu et~al.(2022)Hu, Shen, Wallis, Allen-Zhu, Li, Wang, Wang, and
  Chen]{hu2022lora}
Hu, E.~J., Shen, Y., Wallis, P., Allen-Zhu, Z., Li, Y., Wang, S., Wang, L., and
  Chen, W.
\newblock Lo{RA}: Low-rank adaptation of large language models.
\newblock In \emph{International Conference on Learning Representations}, 2022.

\bibitem[Huang et~al.(2024{\natexlab{a}})Huang, Chen, Wang, Huang, Xu, Wang,
  Liu, Cheng, Zhao, Pang, et~al.]{huang2024chat}
Huang, H., Chen, Y., Wang, Z., Huang, R., Xu, R., Wang, T., Liu, L., Cheng, X.,
  Zhao, Y., Pang, J., et~al.
\newblock Chat-scene: Bridging 3d scene and large language models with object
  identifiers.
\newblock \emph{Advances in Neural Information Processing Systems},
  2024{\natexlab{a}}.

\bibitem[Huang et~al.(2024{\natexlab{b}})Huang, Yong, Ma, Linghu, Li, Wang, Li,
  Zhu, Jia, and Huang]{huang2023embodied}
Huang, J., Yong, S., Ma, X., Linghu, X., Li, P., Wang, Y., Li, Q., Zhu, S.-C.,
  Jia, B., and Huang, S.
\newblock An embodied generalist agent in 3d world.
\newblock \emph{ICML}, 2024{\natexlab{b}}.

\bibitem[Huang et~al.(2025)Huang, Zhang, and
  Tang]{huang20253dr1enhancingreasoning3d}
Huang, T., Zhang, Z., and Tang, H.
\newblock 3d-r1: Enhancing reasoning in 3d vlms for unified scene
  understanding, \emph{arXiv preprint arXiv:2507.23478}, 2025.

\bibitem[Jia et~al.(2024)Jia, Chen, Yu, Wang, Niu, Liu, Li, and
  Huang]{jia2024sceneverse}
Jia, B., Chen, Y., Yu, H., Wang, Y., Niu, X., Liu, T., Li, Q., and Huang, S.
\newblock Sceneverse: Scaling 3d vision-language learning for grounded scene
  understanding.
\newblock In \emph{European Conference on Computer Vision (ECCV)}, 2024.

\bibitem[Johnson et~al.(2017)Johnson, Hariharan, Van Der~Maaten, Hoffman,
  Fei-Fei, Zitnick, and Girshick]{johnson2017inferringexecutingprogramsvisual}
Johnson, J., Hariharan, B., Van Der~Maaten, L., Hoffman, J., Fei-Fei, L.,
  Zitnick, C.~L., and Girshick, R.
\newblock Inferring and executing programs for visual reasoning.
\newblock In \emph{2017 IEEE International Conference on Computer Vision
  (ICCV)}, pp.\  3008--3017. IEEE, 2017.

\bibitem[Jordan et~al.(2024)Jordan, Jin, Boza, You, Cesista, Newhouse, and
  Bernstein]{jordan2024muon}
Jordan, K., Jin, Y., Boza, V., You, J., Cesista, F., Newhouse, L., and
  Bernstein, J.
\newblock Muon: An optimizer for hidden layers in neural networks, 2024.
\newblock URL \url{https://kellerjordan.github.io/posts/muon/}.

\bibitem[Li et~al.(2025{\natexlab{a}})Li, Liu, Guo, Dong, and
  Liu]{Ground_2025_ICRA}
Li, X., Liu, J., Guo, Y., Dong, H., and Liu, Y.
\newblock 3d weakly supervised visual grounding at category and instance
  levels.
\newblock In \emph{Proceedings of the International Conference on Robotics and
  Automation}, 2025{\natexlab{a}}.

\bibitem[Li et~al.(2025{\natexlab{b}})Li, Wang, and Liang]{Li2025r2g}
Li, Y., Wang, Z., and Liang, W.
\newblock R2g: Reasoning to ground in 3d scenes.
\newblock \emph{Pattern Recognition}, 168:\penalty0 111728, 2025{\natexlab{b}}.
\newblock ISSN 0031-3203.

\bibitem[Liu et~al.(2025)Liu, Su, Yao, Jiang, Lai, Du, Qin, Xu, Lu, Yan, Chen,
  Zheng, Liu, Liu, Yin, He, Zhu, Wang, Wang, Dong, Zhang, Kang, Zhang, Xu,
  Zhang, Wu, Zhou, and Yang]{liu2025muonscalablellmtraining}
Liu, J., Su, J., Yao, X., Jiang, Z., Lai, G., Du, Y., Qin, Y., Xu, W., Lu, E.,
  Yan, J., Chen, Y., Zheng, H., Liu, Y., Liu, S., Yin, B., He, W., Zhu, H.,
  Wang, Y., Wang, J., Dong, M., Zhang, Z., Kang, Y., Zhang, H., Xu, X., Zhang,
  Y., Wu, Y., Zhou, X., and Yang, Z.
\newblock Muon is scalable for llm training, \emph{arXiv preprint
  arXiv:2502.16982}, 2025.

\bibitem[Loshchilov et~al.(2019)Loshchilov and Hutter]{loshchilov2017decoupled}
Loshchilov, I. and Hutter, F.
\newblock Decoupled weight decay regularization.
\newblock In \emph{International Conference on Learning Representations}, 2019.

\bibitem[Lyu et~al.(2024)Lyu, Lin, Wang, Yang, Mao, Chen, Xu, Huang, Zhu, Lin,
  et~al.]{mmscan}
Lyu, R., Lin, J., Wang, T., Yang, S., Mao, X., Chen, Y., Xu, R., Huang, H.,
  Zhu, C., Lin, D., et~al.
\newblock Mmscan: A multi-modal 3d scene dataset with hierarchical grounded
  language annotations.
\newblock \emph{Advances in Neural Information Processing Systems},
  37:\penalty0 50898--50924, 2024.

\bibitem[Lyu et~al.(2025)Lyu, Qin, Du, Zhao, and Qiu]{lyu2025}
Lyu, Y., Qin, X., Du, X., Zhao, N., and Qiu, S.
\newblock Multi-path reasoning for multi-hop question answering over knowledge
  graph.
\newblock \emph{Chinese Journal of Electronics}, 34\penalty0 (2):\penalty0
  642--648, 2025.

\bibitem[Ma et~al.(2023)Ma, Yong, Zheng, Li, Liang, Zhu, and
  Huang]{ma2022sqa3d}
Ma, X., Yong, S., Zheng, Z., Li, Q., Liang, Y., Zhu, S.-C., and Huang, S.
\newblock Sqa3d: Situated question answering in 3d scenes.
\newblock In \emph{The Eleventh International Conference on Learning
  Representations}, 2023.

\bibitem[Mao et~al.(2019)Mao, Gan, Kohli, Tenenbaum, and
  Wu]{mao2019neurosymbolicconceptlearnerinterpreting}
Mao, J., Gan, C., Kohli, P., Tenenbaum, J.~B., and Wu, J.
\newblock The neuro-symbolic concept learner: Interpreting scenes, words, and
  sentences from natural supervision.
\newblock In \emph{International Conference on Learning Representations}, 2019.

\bibitem[Mi et~al.(2025)Mi, Wang, Wang, Chen, and Pang]{mi-etal-2025-lasp}
Mi, B., Wang, H., Wang, T., Chen, Y., and Pang, J.
\newblock Language-to-space programming for training-free 3{D} visual
  grounding.
\newblock In \emph{Proceedings of the 2025 Conference on Empirical Methods in
  Natural Language Processing}, pp.\  3844--3864. Association for Computational
  Linguistics, 2025.

\bibitem[Mo et~al.(2024)Mo and Liu]{Mo_Liu_2024}
Mo, W. and Liu, Y.
\newblock Bridging the gap between 2d and 3d visual question answering: A
  fusion approach for 3d vqa.
\newblock \emph{Proceedings of the AAAI Conference on Artificial Intelligence},
  38\penalty0 (5):\penalty0 4261--4268, 2024.

\bibitem[Mo et~al.(2025)Mo, Chen, Peng, Huang, and Liu]{mo2025lego}
Mo, W., Chen, Q., Peng, Y., Huang, S., and Liu, Y.
\newblock Advancing 3d scene understanding with mv-scanqa multi-view reasoning
  evaluation and tripalign pre-training dataset.
\newblock In \emph{Proceedings of the 33rd ACM International Conference on
  Multimedia}, MM '25, pp.\  12973–12980. Association for Computing
  Machinery, 2025.

\bibitem[OpenAI(2023)]{openai2023gpt4}
OpenAI.
\newblock Gpt-4 technical report, \emph{arXiv preprint arXiv:2303.08774}, 2023.

\bibitem[Oquab et~al.(2024)Oquab, Darcet, Moutakanni, Vo, Szafraniec, Khalidov,
  Fernandez, Haziza, Massa, El-Nouby, Assran, Ballas, Galuba, Howes, Huang, Li,
  Misra, Rabbat, Sharma, Synnaeve, Xu, J{\'e}gou, Mairal, Labatut, Joulin, and
  Bojanowski]{Oquab2023DINOv2LR}
Oquab, M., Darcet, T., Moutakanni, T., Vo, H.~Q., Szafraniec, M., Khalidov, V.,
  Fernandez, P., Haziza, D., Massa, F., El-Nouby, A., Assran, M., Ballas, N.,
  Galuba, W., Howes, R., Huang, P.-Y., Li, S.-W., Misra, I., Rabbat, M.,
  Sharma, V., Synnaeve, G., Xu, H., J{\'e}gou, H., Mairal, J., Labatut, P.,
  Joulin, A., and Bojanowski, P.
\newblock Dinov2: Learning robust visual features without supervision.
\newblock \emph{Transactions on Machine Learning Research}, 2024.

\bibitem[Peng et~al.(2025)Peng, Zhang, Zhang, You, Liu, Zhu, Yang, Xu, Geng,
  and Yang]{peng2025lmmr1empowering3blmms}
Peng, Y., Zhang, G., Zhang, M., You, Z., Liu, J., Zhu, Q., Yang, K., Xu, X.,
  Geng, X., and Yang, X.
\newblock Lmm-r1: Empowering 3b lmms with strong reasoning abilities through
  two-stage rule-based rl, \emph{arXiv preprint arXiv:2503.07536}, 2025.

\bibitem[Peng et~al.(2026)Peng, Wang, Li, Zheng, Yin, and He]{peng2025}
Peng, Y., Wang, Z., Li, G., Zheng, X., Yin, S., and He, H.
\newblock A survey on fine-grained multimodal large language models.
\newblock \emph{Chinese Journal of Electronics}, 35\penalty0 (2):\penalty0
  771--803, 2026.

\bibitem[Qin et~al.(2025)Qin, Xu, and Liu]{Qin2025ApplyHT}
Qin, Y., Xu, Z., and Liu, Y.
\newblock Apply hierarchical-chain-of-generation to complex attributes
  text-to-3d generation.
\newblock In \emph{Proceedings of the IEEE/CVF Conference on Computer Vision
  and Pattern Recognition (CVPR)}, pp.\  18521--18530, 2025.

\bibitem[Qu et~al.(2026)Qu, Li, Chen, Liu, Shi, Ren, Jing, and
  Zhang]{qu2025segdino3d3dinstancesegmentation}
Qu, J., Li, H., Chen, X., Liu, S., Shi, Y., Ren, T., Jing, R., and Zhang, L.
\newblock Segdino3d: 3d instance segmentation empowered by both image-level and
  object-level 2d features.
\newblock In \emph{Proceedings of the AAAI Conference on Artificial
  Intelligence}, volume~40, pp.\  8649--8658, 2026.

\bibitem[Schult et~al.(2023)Schult, Engelmann, Hermans, Litany, Tang, and
  Leibe]{Schult23ICRA}
Schult, J., Engelmann, F., Hermans, A., Litany, O., Tang, S., and Leibe, B.
\newblock {Mask3D: Mask Transformer for 3D Semantic Instance Segmentation}.
\newblock In \emph{{International Conference on Robotics and Automation
  (ICRA)}}, 2023.

\bibitem[Shao et~al.(2024)Shao, Wang, Zhu, Xu, Song, Bi, Zhang, Zhang, Li, Wu,
  and Guo]{shao2024deepseekmathpushinglimitsmathematical}
Shao, Z., Wang, P., Zhu, Q., Xu, R., Song, J., Bi, X., Zhang, H., Zhang, M.,
  Li, Y.~K., Wu, Y., and Guo, D.
\newblock Deepseekmath: Pushing the limits of mathematical reasoning in open
  language models, \emph{arXiv preprint arXiv:2402.03300}, 2024.

\bibitem[Vedantam et~al.(2019)Vedantam, Desai, Lee, Rohrbach, Batra, and
  Parikh]{vedantam2019probabilisticneuralsymbolicmodelsinterpretable}
Vedantam, R., Desai, K., Lee, S., Rohrbach, M., Batra, D., and Parikh, D.
\newblock Probabilistic neural symbolic models for interpretable visual
  question answering.
\newblock In \emph{Proceedings of the 36th International Conference on Machine
  Learning}, volume~97 of \emph{Proceedings of Machine Learning Research}, pp.\
   6428--6437. PMLR, 2019.

\bibitem[Yang et~al.(2025)Yang, Li, Yang, Zhang, Hui, Zheng, Yu, Gao, Huang,
  Lv, Zheng, Liu, Zhou, Huang, Hu, Ge, Wei, Lin, Tang, Yang, Tu, Zhang, Yang,
  Yang, Zhou, Zhou, Lin, Dang, Bao, Yang, Yu, Deng, Li, Xue, Li, Zhang, Wang,
  Zhu, Men, Gao, Liu, Luo, Li, Tang, Yin, Ren, Wang, Zhang, Ren, Fan, Su,
  Zhang, Zhang, Wan, Liu, Wang, Cui, Zhang, Zhou, and
  Qiu]{yang2025qwen3technicalreport}
Yang, A., Li, A., Yang, B., Zhang, B., Hui, B., Zheng, B., Yu, B., Gao, C.,
  Huang, C., Lv, C., Zheng, C., Liu, D., Zhou, F., Huang, F., Hu, F., Ge, H.,
  Wei, H., Lin, H., Tang, J., Yang, J., Tu, J., Zhang, J., Yang, J., Yang, J.,
  Zhou, J., Zhou, J., Lin, J., Dang, K., Bao, K., Yang, K., Yu, L., Deng, L.,
  Li, M., Xue, M., Li, M., Zhang, P., Wang, P., Zhu, Q., Men, R., Gao, R., Liu,
  S., Luo, S., Li, T., Tang, T., Yin, W., Ren, X., Wang, X., Zhang, X., Ren,
  X., Fan, Y., Su, Y., Zhang, Y., Zhang, Y., Wan, Y., Liu, Y., Wang, Z., Cui,
  Z., Zhang, Z., Zhou, Z., and Qiu, Z.
\newblock Qwen3 technical report, \emph{arXiv preprint arXiv:2505.09388}, 2025.

\bibitem[Yi et~al.(2018)Yi, Wu, Gan, Torralba, Kohli, and
  Tenenbaum]{Yi2018NeuralSymbolicVD}
Yi, K., Wu, J., Gan, C., Torralba, A., Kohli, P., and Tenenbaum, J.~B.
\newblock Neural-symbolic vqa: Disentangling reasoning from vision and language
  understanding.
\newblock In \emph{Neural Information Processing Systems}, 2018.

\bibitem[Yu et~al.(2025)Yu, Li, Wang, Chen, and Zhu]{Inst3D-LMM}
Yu, H., Li, W., Wang, S., Chen, J., and Zhu, J.
\newblock Inst3d-lmm: Instance-aware 3d scene understanding with multi-modal
  instruction tuning.
\newblock In \emph{Proceedings of the IEEE/CVF Conference on Computer Vision
  and Pattern Recognition (CVPR)}, pp.\  14147--14157, 2025.

\bibitem[Yuan et~al.(2025)Yuan, Jiang, Feng, Zhang, Cui, Li, and
  Zhao]{yuan2025scener1videogroundedlargelanguage}
Yuan, Z., Jiang, S., Feng, C.-M., Zhang, Y., Cui, S., Li, Z., and Zhao, N.
\newblock Scene-r1: Video-grounded large language models for 3d scene reasoning
  without 3d annotations, \emph{arXiv preprint arXiv:2506.17545}, 2025.

\bibitem[Zhang et~al.(2025)Zhang, Huang, Yao, Liu, Zhang, Lu, and
  Tao]{zhang2025r1}
Zhang, J., Huang, J., Yao, H., Liu, S., Zhang, X., Lu, S., and Tao, D.
\newblock R1-vl: Learning to reason with multimodal large language models via
  step-wise group relative policy optimization.
\newblock In \emph{Proceedings of the IEEE/CVF International Conference on
  Computer Vision (ICCV)}, pp.\  1859--1869, 2025.

\bibitem[Zhang et~al.(2023)Zhang, Gong, and Chang]{zhang2023multi3drefer}
Zhang, Y., Gong, Z., and Chang, A.~X.
\newblock Multi3drefer: Grounding text description to multiple 3d objects.
\newblock In \emph{Proceedings of the IEEE/CVF International Conference on
  Computer Vision (ICCV)}, pp.\  15225--15236, 2023.

\bibitem[Zhao et~al.(2025)Zhao, Xu, Chen, Peng, and
  Liu]{zhao2025domaingaps-arxiv}
Zhao, Z., Xu, Z., Chen, Q., Peng, Y., and Liu, Y.
\newblock Investigating domain gaps for indoor 3d object detection.
\newblock In \emph{Proceedings of the 33rd ACM International Conference on
  Multimedia}, MM '25, pp.\  13198–13205. Association for Computing
  Machinery, 2025.

\bibitem[Zheng et~al.(2025)Zheng, Huang, and
  Wang]{zheng2024video3dllmlearningpositionaware}
Zheng, D., Huang, S., and Wang, L.
\newblock Video-3d llm: Learning position-aware video representation for 3d
  scene understanding.
\newblock In \emph{Proceedings of the IEEE/CVF Conference on Computer Vision
  and Pattern Recognition (CVPR)}, pp.\  8995--9006, 2025.

\bibitem[Zheng et~al.(2024)Zheng, Yin, Xie, Sun, Huang, Yu, Cao, Kozyrakis,
  Stoica, Gonzalez, et~al.]{zheng2024sglangefficientexecutionstructured}
Zheng, L., Yin, L., Xie, Z., Sun, C., Huang, J., Yu, C.~H., Cao, S., Kozyrakis,
  C., Stoica, I., Gonzalez, J.~E., et~al.
\newblock Sglang: Efficient execution of structured language model programs.
\newblock In \emph{Advances in Neural Information Processing Systems},
  volume~37, pp.\  62557--62583, 2024.

\bibitem[Zhou et~al.(2024)Zhou, Wang, Ma, Liu, Huang, and
  Wang]{Zhou2023Uni3DEU}
Zhou, J., Wang, J., Ma, B., Liu, Y.-S., Huang, T., and Wang, X.
\newblock Uni3d: Exploring unified 3d representation at scale.
\newblock In \emph{International Conference on Learning Representations}, 2024.

\bibitem[Zhou et~al.(2025)Zhou, Liu, and Zheng]{zhou2025aqua}
Zhou, S., Liu, Y., and Zheng, F.
\newblock Learn 3d vqa better with active selection and reannotation.
\newblock In \emph{Proceedings of the 33rd ACM International Conference on
  Multimedia}, MM '25, pp.\  4610–4618. Association for Computing Machinery,
  2025.

\bibitem[Zhou et~al.(2026)Zhou, Zheng, Zheng, and
  Liu]{zhou2026scalableobjectrelationencoding}
Zhou, S., Zheng, M., Zheng, F., and Liu, Y.
\newblock Scalable object relation encoding for better 3d spatial reasoning in
  large language models, \emph{arXiv preprint arXiv:2603.24721}, 2026.

\bibitem[Zhu et~al.(2025)Zhu, Wang, Zhang, Pang, and
  Liu]{zhu2025llava3dsimpleeffectivepathway}
Zhu, C., Wang, T., Zhang, W., Pang, J., and Liu, X.
\newblock Llava-3d: A simple yet effective pathway to empowering lmms with 3d
  capabilities.
\newblock In \emph{Proceedings of the IEEE/CVF International Conference on
  Computer Vision (ICCV)}, pp.\  4295--4305, 2025.

\bibitem[Zhu et~al.(2023)Zhu, Ma, Chen, Deng, Huang, and Li]{zhu20233dvista}
Zhu, Z., Ma, X., Chen, Y., Deng, Z., Huang, S., and Li, Q.
\newblock 3d-vista: Pre-trained transformer for 3d vision and text alignment.
\newblock In \emph{Proceedings of the IEEE/CVF International Conference on
  Computer Vision (ICCV)}, pp.\  2911--2921, 2023.

\bibitem[Zhu et~al.(2024)Zhu, Zhang, Ma, Niu, Chen, Jia, Deng, Huang, and
  Li]{zhu2024pq3d}
Zhu, Z., Zhang, Z., Ma, X., Niu, X., Chen, Y., Jia, B., Deng, Z., Huang, S.,
  and Li, Q.
\newblock Unifying 3d vision-language understanding via promptable queries.
\newblock In \emph{European Conference on Computer Vision (ECCV)}, pp.\
  188–206. Springer-Verlag, 2024.

\end{thebibliography}
\bibliographystyle{icml2026_singleauthoronly}

\newpage
\appendix
\onecolumn
\section{Supplementary Material}
\label{sec:appendix}
In this supplementary material, we provide comprehensive details to support reproducibility and further understanding of our method.
\Cref{sec:impl_details} presents implementation details including detailed model architecture specifications and training protocols for all three curriculum stages.
\Cref{sec:data_details} elaborates on the data construction process: the perception alignment mixture (Stage 1), the program-to-CoT translation pipeline with verified execution traces (Stage 2), and the RL training configuration (Stage 3).
\Cref{tab:prompts} and \Cref{fig:full_prompt_example} provide the instruction templates and a complete reasoning trace example used throughout training and evaluation.
\Cref{sec:additional_exp} presents additional experiments on backbone scaling and reasoning trace quality analysis.
\Cref{sec:modular_example} further explains how modular inference enhancement works with concrete examples.
Finally, \Cref{sec:limitations} discusses limitations of our approach.

\subsection{Implementation Details}
\label{sec:impl_details}

\paragraph{Model Architecture}
We implement \methodName{} on Qwen3-VL-Instruct~\cite{yang2025qwen3technicalreport}, a state-of-the-art 8B parameter multi-modal large language model. For 3D scene encoding, we follow prior work~\cite{huang2024chat} to segment input scenes using Mask3D~\cite{Schult23ICRA} and extract visual features using Uni3D~\cite{Zhou2023Uni3DEU} for 3D geometric information and DINOv2~\cite{Oquab2023DINOv2LR} for visual appearance features.

For object spatial encoding, we adopt a lattice-based positional encoding that represents each object's 6D configuration (location $x, y, z$ and size $h, w, l$) as a continuous embedding.
Specifically, for each spatial axis $a \in \{x, y, z, h, w, l\}$, we define a learnable embedding matrix $\bm{H}^{a} \in \mathbb{R}^{N \times d}$, where $N=200$ is the number of uniformly spaced grid points along the axis.
Given an object with normalized configuration $\bm{p} = (x, y, z, h, w, l) \in [0,1]^6$, we compute the positional embedding as:
\begin{equation}
    \bm{h}_{\text{pos}}(\bm{p}) = \sum_{a \in \{x,y,z,h,w,l\}} \sum_{k=1}^{N} w_k^{a} \cdot \bm{H}^{a}_k, \quad \text{where} \quad w_k^{a} = \exp\left(-\left(N \cdot p_a - k\right)^2\right)
\end{equation}
Intuitively, the Gaussian weights $w_k^a$ assign higher influence to grid points closer to the object's actual coordinate, producing a smooth interpolation over the learnable lattice embeddings.
The final object representation concatenates this spatial embedding with visual features and is interleaved with instruction tokens as LLM input.

\paragraph{Training Protocol}
We employ LoRA~\cite{hu2022lora} with rank $r=128$ for parameter-efficient fine-tuning across all stages. Stage 1 (Perception Alignment) trains for 3 epochs on our alignment data mixture with a learning rate of $1 \times 10^{-5}$ and batch size of 16. Stage 2 (CoT-SFT) trains for 1 epoch on the program-derived reasoning traces with a learning rate of $1 \times 10^{-5}$ and batch size of 8. Stage 3 (CoT-RL) employs Group Relative Policy Optimization (GRPO)~\cite{shao2024deepseekmathpushinglimitsmathematical} with a group size of $G=8$ rollouts per prompt, trained for 1 epoch with a learning rate of $3 \times 10^{-6}$ with a batch size of 128 prompts (1024 rollouts). We set $\varepsilon=0.2$ for GRPO clipping and $\beta=0$ as KL divergence coefficient to further reduce GPU memory consumption and enhance exploration.
All experiments use AdamW~\cite{loshchilov2017decoupled} optimizer with weight decay of 0.001 and are conducted on 8 NVIDIA A100-40G GPUs.

A significant engineering challenge for Stage 3 is the lack of native 3D multi-modal support in existing RL libraries. To address this, we developed a custom infrastructure built upon SGLang~\cite{zheng2024sglangefficientexecutionstructured} that enables efficient policy rollouts with arbitrary 3D object-centric multi-modal features. Our implementation combines acceleration techniques from SGLang and achieves approximately 3$\times$ speedup compared to naive generation with huggingface's \texttt{transformers} library, making large-scale 3D RL exploration and training tractable.

\subsection{Data Construction Details}
\label{sec:data_details}

\paragraph{Stage 1: Perception Alignment}
To establish robust 3D scene understanding capabilities, we first construct a balanced mixture of object-centric tasks. These tasks include
\begin{itemize}
    \item \textbf{Object Recognition:} Tasks include category ("What is this object?") and attribute ("What attribute does this object have?") identification, targeting only single-object feature inputs. We collect object category annotations from ScanNet~\cite{dai2017scannet} and attribute annotations from MMScan~\cite{mmscan}.
    Sourced from ScanNet and textual attributes extracted from captions.
    \item \textbf{Object Localization:} Tasks involve outputting 3D coordinates of specified objects based on its object ID ("Where is the object with ID \texttt{<id>} located?"). We utilize ground-truth bounding box centers from ScanNet~\cite{dai2017scannet} as supervision.
    \item \textbf{Object Captioning:} Generating one-sentence descriptions for single objects using ScanRefer~\cite{chen2020scanrefer} and ReferIt3D~\cite{achlioptas2020referit_3d} captions where the target is unambiguous.
\end{itemize}
The total alignment dataset comprises approximately 193K instruction-response pairs.

\paragraph{Stage 2: Symbolic Reasoning Injection via CoT-SFT}
For CoT-SFT, we translate neuro-symbolic programs into iterative natural language reasoning traces. Each program step is converted into a \textbf{Plan} sentence describing the operation, and an \textbf{Execution} block detailing the intermediate results. We build the final CoT by first concatenating all step-by-step Plans and then all Executions, followed by a simple summarization into the final answer.
To enable thinking, we simply add a sentence "Think about the scene first." in the instruction prompt sent to model during CoT-SFT, which is also perserved at CoT-RL stage.

For Level 1 programs, we construct single-step \texttt{filter} operations from ScanNet category annotations and MMScan attribute annotations, yielding approximately 78K traces with verified outputs. For Level 2 programs, we leverage the Sr3D dataset~\cite{achlioptas2020referit_3d}, which provides synthetic instructions guaranteed to be solvable in exactly two steps. Each instruction maps to a two-step program (e.g., \texttt{relate(filter(A), filter(B), r)}), and we construct complete execution traces by verifying intermediate \texttt{filter} outputs against object annotations and final \texttt{relate} outputs against target annotations. This yields approximately 66K traces with perfect step-by-step supervision. The specific templates used are as follows:
\begin{itemize}
    \item \textbf{Plan Generation:}
    \begin{itemize}
        \item \texttt{filter(C)} $\to$ "Find all objects of category [C]."
        \item \texttt{relate(A, B, r)} $\to$ "Check which [A] are [r] to [B]."
    \end{itemize}
    \item \textbf{Execution Generation:}
    \begin{itemize}
        \item \texttt{scene()} $\to$ I see [$N$] objects in the scene: Object 1: [Category\textsubscript{1}], Object 2: [Category\textsubscript{2}], ...
        \item \texttt{filter()} $\to$ Looking for [Category]... I found [$k$] objects: Object [ID\textsubscript{1}]: at [Position], Object [ID\textsubscript{2}]: at [Position], ...
        \item \texttt{relate()} $\to$ Analyzing relation [r]... Object [ID\textsubscript{A}] is [r] to Object [ID\textsubscript{B}], Object [ID\textsubscript{C}] is [not r] to Object [ID\textsubscript{D}]...
    \end{itemize}
\end{itemize}
The full algorithm for building CoT data from programs is summarized in Algorithm~\ref{alg:build_cot}, and a full example of the assembled CoT is shown in \Cref{fig:full_prompt_example}.

\begin{algorithm}[!ht]
\small
\setlength{\algomargin}{0.5em}
\SetNlSty{texttt}{}{:}  
\SetInd{0.3em}{0.6em}

\SetAlgoLined
\DontPrintSemicolon
\KwData{Program $P$, Ground truth annotations $\mathcal{A}$}
\KwResult{Chain-of-Thought data with plans $\mathbf{Plans}$ and executions $\mathbf{Execs}$}

\tcp{Parse program into execution sequence}
$\mathcal{T} \gets \textsc{ParseAST}(P)$\;
$\mathcal{S} \gets \textsc{GetExecutionSequence}(\mathcal{T})$\;

\tcp{Initialize plan and execution lists}
$\mathbf{Plans} \gets [\,]$\;
$\mathbf{Execs} \gets [\,]$\;

\tcp{Process each execution step}
\ForEach{statement $s$ in $\mathcal{S}$}{
    \uIf{$s$ is \texttt{scene()}}{
        $\mathcal{O}_{in} \gets \emptyset$\;
        $\mathcal{O}_{out} \gets \text{all objects in scene}$\;
    }
    \uElseIf{$s$ is \texttt{filter(condition)}}{
        $\mathcal{O}_{in} \gets \mathcal{O}$\;
        $\mathcal{O}_{out} \gets \{o \in \mathcal{O} \mid o \text{ matches } condition\}$\;
    }
    \uElseIf{$s$ is \texttt{relate(...)} or \texttt{relate\_anchor(...)} or \texttt{relate\_triple(...)}}{
        $\mathcal{O}_{in} \gets$ filtered object sets from previous steps\;
        $\mathcal{O}_{out} \gets$ target objects from $\mathcal{A}$\;
    }

    \tcp{Generate plan and execution for this step}
    $plan \gets \textsc{ProgramToPlan}(s)$\;
    $exec \gets \textsc{FormatExecution}(s,\mathcal{O}_{in}, \mathcal{O}_{out})$\;

    $\mathbf{Plans}.\textsc{append}(plan)$\;
    $\mathbf{Execs}.\textsc{append}(exec)$\;
}

\tcp{Concatenate plans and executions}
$CoT_{plans} \gets \textsc{Concatenate}(\mathbf{Plans})$\;
$CoT_{execs} \gets \textsc{Concatenate}(\mathbf{Execs})$\;

\Return{$\textsc{Concatenate}(CoT_{plans}, CoT_{execs})$}\;

\caption{Building CoT Data from Programs}
\label{alg:build_cot}
\end{algorithm}

\paragraph{Stage 3: Open-Set and Complex Reasoning Generalization via CoT-RL}
For reinforcement learning, we use the training splits of ScanRefer~\cite{chen2020scanrefer} and Multi3DRefer~\cite{zhang2023multi3drefer}, which contain natural language instructions with complex nested structures and open-vocabulary concepts. Since ground-truth execution traces are unavailable for these datasets, we rely solely on outcome supervision (whether the predicted bounding box matches the target) combined with our format reward to guide exploration.

\subsection{Instruction Prompts}
\Cref{tab:prompts} summarizes the instruction and response templates used in \methodName{}. In scene context prompts, \texttt{|object\_set|} placeholder will be replaced by real object-centric scene features. Object positions and dimensions are formatted with 2 decimal places precision.

\begin{table*}[!ht]
\centering
\caption{\textbf{Prompt Templates.} Representative templates for scene context, instructions, object formatting, and responses used in \methodName{}.}
\label{tab:prompts}
\small
\begin{tabularx}{\linewidth}{l X}
\toprule
\textbf{Component} & \textbf{Template} \\
\midrule
Scene Context & \texttt{These are all objects in the scene: |object\_set|}  \\
\midrule
Instruction & \texttt{Think about the scene first. Identify the object described as: "\{description\}". Respond with the object's ID, position, and size.} \\
\midrule
Object Locations & \texttt{Object \{id\}: \{name\} at (\{x\}, \{y\}, \{z\}), size: \{w\} x \{h\} x \{d\}} \\
\midrule
Response & \texttt{I found \{count\} object(s) matching the description: \{object\_locations\}} \\
\midrule
Chain-of-Thoughts & \texttt{[think] \{thinking procedure...\} [answer]} \\
\bottomrule
\end{tabularx}
\vspace{2em}
\end{table*}

\begin{figure}[!ht]
\centering
\begin{promptbox}
These are all objects in the scene: |object\_set|.
Think about the scene first. Find the vase left to the computer. Provide its ID, position, and dimensions.

\think{[think]}

I need to find the vase left to the computer.

\commenttt{// Stepwise plans.}

Let's plan my next steps: Examine all objects in the scene; Find all objects of category 'vase'; Find all objects of category 'computer'; Check which vase(s) are 'left to' the computer.

\commenttt{// Execution of scene() statement.}

First, let me list the object IDs and names:
Object 0(vase), 1(vase), 2(bottle), 3(rug), ... \textcolor{gray}{[others omitted]}

\commenttt{// Execution of filter() statement.}

Among these objects, I can see the vase(s):

Object 0: At (1.15, 6.09, 1.33), size: 0.86 x 0.99 x 1.79

Object 1: At (6.66, 5.41, 0.29), size: 1.60 x 1.09 x 1.70

And the computer(s):

Object 4: At (2.34, 3.50, 0.75), size: 1.20 x 0.60 x 1.10

Object 11: At (1.44, 4.04, 3.10), size: 0.87 x 1.38 x 0.87

\commenttt{// Execution of relate() statement.}

Now, I will analyze if any vase is left to the computer:

The vase (ID 0) is left to the computer (ID 4).

The vase (ID 0) is above and near the computer (ID 11).

The vase (ID 1) is right to and near the computer (ID 4).

The vase (ID 1) is right to, in front of and far from the computer (ID 11).

\commenttt{// Summarization.}

From analysis above, the vase (ID 0) is left to a computer (ID 4).
Now, I will formulate the response based on the identified objects.

\answer{[answer]}

I found the vase left to the computer:
Object 0: At (1.15, 6.09, 1.33), size: 0.86 x 0.99 x 1.79
\end{promptbox}
\caption{A full example of assembled thinking trace for \methodName{}. The trace includes the scene context, instruction, step-by-step thinking trace with plans and executions, and the final answer.}
\label{fig:full_prompt_example}
\end{figure}

\subsection{Additional Experiments}
\label{sec:additional_exp}

\paragraph{Backbone Scaling}
To evaluate the impact of the underlying MLLM backbone size on 3D spatial reasoning performance, we conduct experiments with different model scales within our \methodName{} framework. As shown in Table~\ref{tab:backbone_scaling}, scaling up from 4B to 8B parameters yields consistent performance improvements across all metrics, demonstrating the benefits of larger model capacity for 3D vision-language understanding.

\begin{table*}[htbp]
\centering
\caption{\textbf{Backbone Scaling.} Performance comparison of \methodName{} with different backbone sizes. Larger backbones yield consistent improvements.}
\label{tab:backbone_scaling}
\small
\begin{tabularx}{\linewidth}{l YY YY}
\toprule
\multirow{2}{*}{\textbf{Backbone}} & \multicolumn{2}{c}{\textbf{ScanRefer}} & \multicolumn{2}{c}{\textbf{Multi3DRefer}} \\
\cmidrule(lr){2-3} \cmidrule(lr){4-5}
& Acc@0.25 & Acc@0.5 & F1@0.25 & F1@0.5 \\
\midrule
Qwen3-VL-4B  & 56.4 & 48.9 & 56.1 & 51.0 \\
Qwen3-VL-8B  & \textbf{58.4} & \textbf{51.2} & \textbf{59.2} & \textbf{53.8} \\
\midrule
\textcolor{ForestGreen}{\(\Delta\) (4B \(\to\) 8B)} & \textcolor{ForestGreen}{+2.0} & \textcolor{ForestGreen}{+2.3} & \textcolor{ForestGreen}{+3.1} & \textcolor{ForestGreen}{+2.8} \\
\bottomrule
\end{tabularx}
\vspace{1em}
\end{table*}

\paragraph{Reasoning-Trace Quality Analysis}
A natural concern is whether outcome-RL in Stage~3 (CoT-RL Stage) preserves the structured reasoning patterns injected in Stage~2 (CoT-SFT), or degrades them into reward-hacking shortcuts. We then conduct a systematic quality analysis on 200 randomly sampled reasoning traces from the ScanRefer validation set after Stage~3 training.

Specifically, we use Claude Opus 4.6~\cite{opus4_5} as an automated judge, evaluating each trace along two criteria:
\begin{itemize}[nosep,leftmargin=1.2em]
  \item \textbf{Neuro-Symbolic (NS-) Compliance:}
    Whether the trace is parsable into a valid symbolic program structure (i.e., a well-formed sequential combination of primitives like \texttt{scene}, \texttt{filter} and \texttt{relate}).
  \item \textbf{Reference Consistency:}
    Whether objects are consistently referenced (same ID $\leftrightarrow$ same entity) throughout the chain, without contradictory or hallucinated references.
\end{itemize}
We validate the automated judge on 50 manually annotated traces
and observe \textbf{94\% agreement} with human annotations, confirming its reliability.

As shown in \Cref{tab:trace_quality}, two findings emerge.
\textbf{(a)}~RL preserves structured reasoning: 92.5\% of post-RL traces remain parsable into symbolic programs, demonstrating that outcome-only RL does not collapse the reasoning structure injected in Stage~2, even though the predicates are now open-vocabulary natural language that goes beyond what traditional NS3D frameworks can express.
\textbf{(b)}~Reasoning quality correlates with answer correctness: correctly grounded samples show \textbf{+13.4\%} higher joint compliance than incorrect ones, indicating that the model's grounding accuracy is linked to its reasoning quality rather than memorized shortcuts.
These results confirm that our curriculum could successfully internalize faithful, verifiable reasoning patterns that persist through RL optimization and directly contribute to downstream performance.

\begin{table}[ht]
\centering
\caption{%
  \textbf{Reasoning-trace quality analysis}.
  Correctly grounded samples exhibit substantially higher structural compliance than incorrect ones, indicating that structured reasoning directly contributes to accuracy.
}
\label{tab:trace_quality}
\small
\begin{tabularx}{\linewidth}{l YYYY}
\toprule
\textbf{Criterion}
  & \textbf{Overall}
  & \textbf{In Correct Samples}
  & \textbf{In Incorrect Samples}
  & \textbf{$\Delta$} \\
\midrule
NS-Compliance
  & 92.5 & 95.3 & 90.4
  & \textcolor{ForestGreen}{+4.9} \\
Reference Consistency
  & 86.5 & 91.8 & 82.6
  & \textcolor{ForestGreen}{+9.2} \\
\midrule
Both criteria met
  & 80.5 & 88.2 & 74.8
  & \textcolor{ForestGreen}{+13.4} \\
\bottomrule
\end{tabularx}
\vspace{1em}
\end{table}

\subsection{Modular Inference Enhancement Example}
\label{sec:modular_example}

\Cref{fig:full_prompt_example} illustrates a complete trace where the model generates all four thinking phases (planning, scene perception, execution, answer) end-to-end without external modules.
We further provide a concrete example to demonstrate the modular replacement mechanism described in \Cref{sec:modularity}, based on an examplar query \textit{``find the chair to the left of the desk near the window.''}

\paragraph{External Plan Injection} As shown in \Cref{fig:modular_plan}, the Planning block can be replaced by an external planner (e.g., Claude Opus 4.5~\cite{opus4_5}) before decoding begins. The model then decodes all subsequent phases normally while conditioned on the injected plan.

\paragraph{Perception Module Replacement} \Cref{fig:modular_scene} further illustrates the case where the model decodes the Planning block normally, after which the \texttt{scene()} output is replaced by a stronger 3D perception model (e.g., SegDINO3D~\cite{qu2025segdino3d3dinstancesegmentation}).
The model continues decoding and benefit from the more accurate object list.

In both replacement cases (\Cref{fig:modular_plan,fig:modular_scene}), the model simply conditions on the injected text as if it had generated it itself, and continues decoding the remaining phases without any modification or retraining.

\begin{figure}[!h]
\centering

\begin{subfigure}{\linewidth}
\centering
\begin{promptbox}
\small
\think{[think]}

I need to find: ``the chair to the left of the desk near the window.''

\commenttt{// Planning \textcolor{blue}{(E.g., injected from Claude Opus 4.5)}}

\textcolor{blue}{Let's plan: First find all `window'; Then find `desk' that are `near' those windows; Finally find `chair' that are `to the left of' those desks.}

\commenttt{// Execute scene, filter, relate ...}

\textit{(model decodes normally from here)} ...

\answer{[answer]}

\textit{(model answers normally)}
\end{promptbox}
\caption{External plan injection. \textcolor{blue}{Blue text} is injected from an external planner; the model continues decoding the remaining phases.}
\label{fig:modular_plan}
\end{subfigure}

\vspace{0.8em}

\begin{subfigure}{\linewidth}
\centering
\begin{promptbox}
\small
\think{[think]}

I need to find: ``the chair to the left of the desk near the window.''

\commenttt{// Planning}

Let's plan: Find objects of category `chair', `desk', and `window'; ...

\textit{(model decodes normally up to here)}

\commenttt{// Execute scene() \textcolor{ForestGreen}{(E.g., injected from SegDINO3D)}}

\textcolor{ForestGreen}{I see 33 object(s) in the scene: 0 (sofa), 1 (armchair), 2 (desk), 3 (window), 4 (floor lamp), ...} \textcolor{gray}{[more objects omitted]}

\commenttt{// Execute filter, relate ...}

\textit{(model decodes normally, benefiting from more accurate perception)} ...

\answer{[answer]}

\textit{(model answers normally)}
\end{promptbox}
\caption{Perception module replacement. \textcolor{ForestGreen}{Green text} is injected from an external detector; the model continues decoding after the enhanced perception.}
\label{fig:modular_scene}
\end{subfigure}

\caption{Modular inference enhancement examples.}
\label{fig:modular_examples}
\end{figure}

\subsection{Limitations}
\label{sec:limitations}

While \methodName{} demonstrates strong performance on 3D spatial reasoning benchmarks, several limitations merit discussion.

\noindent\textbf{Perception Bottleneck.} As evidenced by our modularity analysis (\textbf{Modularity and Scalability.} in~\Cref{subsec:ablation}), the system's overall performance remains fundamentally bounded by the accuracy of the underlying 3D perception module; when object detection or semantic recognition fails, the reasoning engine cannot recover regardless of its logical correctness, as it operates on incomplete or erroneous scene representations.

\noindent\textbf{Deep Reasoning Chains.} Although our method exhibits emergent composition of logical primitives such as \texttt{intersection} and \texttt{union} for multi-condition constraints, the reliability and coherence of very deep reasoning chains involving five or more nested operations have not been extensively validated, and such complex traces may be susceptible to error accumulation during RL optimization and is rare in current datasets.

\noindent\textbf{Physics and Dynamics Reasoning.}
Our symbolic schema targets compositional spatial reasoning over static scene geometry. Queries that require understanding of physical dynamics or causal simulation (e.g., \textit{``which supporting furniture would cause the shelf to collapse if removed?''}) fall outside the current primitive set, as they demand commonsense reasoning about forces, stability, and temporal state changes that are complementary to the spatial decomposition our framework provides. Extending the primitive vocabulary to include physics-aware predicates is a promising future direction.

\noindent\textbf{Generalization to Outdoor and Navigation Domains.} Our experiments focus exclusively on indoor scenes from ScanNet-family benchmarks. While the core reasoning primitives (\texttt{scene}, \texttt{filter}, \texttt{relate}) are domain-general in principle, we have not empirically validated this transfer. Domain adaptation would likely require extending the primitive set with domain-specific predicates (e.g., traversability, agent dynamics) and constructing seed program traces from outdoor datasets with object pose annotations. We consider this a valuable direction for future work.

\end{document}